\def\year{2022}\relax
\documentclass[letterpaper]{article} 
\usepackage{aaai22}  
\usepackage{times}  
\usepackage{helvet}  
\usepackage{courier}  
\usepackage[hyphens]{url}  
\usepackage{graphicx} 
\def\UrlFont{\rm}  
\usepackage{natbib}  
\usepackage{caption} 
\DeclareCaptionStyle{ruled}{labelfont=normalfont,labelsep=colon,strut=off} 
\usepackage{xspace} 
\usepackage{multirow}
\usepackage{xcolor}
\usepackage{comment}
\usepackage{subcaption}
\usepackage{enumerate}
\usepackage{enumitem}
\usepackage{amsmath}
\usepackage{amsfonts}
\usepackage{amsthm}
\usepackage[english]{babel}
\usepackage[switch]{lineno}
\usepackage{xspace}
\usepackage{wasysym}
\usepackage{url}

\newcommand{\ORplayer}{\text{$\pi_{\square}$}\xspace}
\newcommand{\ANDplayer}{\text{$\pi_{\ocircle}$}\xspace}

\usepackage{algorithm}
\usepackage[noend]{algpseudocode}

\algrenewcommand\algorithmicforall{\textbf{foreach}}

\newtheorem{lemma}{Lemma}

\begin{document}
\title{A Novel Approach to Solving Goal-Achieving Problems for Board Games}
\author{
Chung-Chin Shih\textsuperscript{\rm 1,2}\equalcontrib, 
Ti-Rong Wu\textsuperscript{\rm 1}\equalcontrib, Ting Han Wei\textsuperscript{\rm 3}, and I-Chen Wu\textsuperscript{\rm 1,2}\thanks{Corresponding author.}
}
\affiliations{
    \textsuperscript{\rm 1}Department of Computer Science, National Yang Ming Chiao Tung University, Hsinchu, Taiwan\\
    \textsuperscript{\rm 2}Research Center for Information Technology Innovation, Academia Sinica, Taipei, Taiwan\\
    \textsuperscript{\rm 3}Department of Computing Science, University of Alberta, Edmonton, Canada\\
    \{rockmanray,kds285\}@aigames.nctu.edu.tw, tinghan@ualberta.ca, icwu@cs.nctu.edu.tw 


}

\maketitle

\begin{abstract}
\emph{Goal-achieving} problems are puzzles that set up a specific situation with a clear objective.
An example that is well-studied is the category of \emph{life-and-death} (L\&D) problems for Go, which helps players hone their skill of identifying region safety. 
Many previous methods like lambda search try null moves first, then derive so-called relevance zones (RZs), outside of which the opponent does not need to search.
This paper first proposes a novel RZ-based approach, called the {\em RZ-Based Search (RZS)}, to solving L\&D problems for Go. 
RZS tries moves before determining whether they are null moves post-hoc.
This means we do not need to rely on null move heuristics, resulting in a more elegant algorithm, so that it can also be seamlessly incorporated into AlphaZero's super-human level play in our solver.
To repurpose AlphaZero for solving, we also propose a new training method called {\em Faster to Life (FTL)}, which modifies AlphaZero to entice it to win more quickly.
We use RZS and FTL to solve L\&D problems on Go, namely solving 68 among 106 problems from a professional L\&D book while a previous state-of-the-art program TSUMEGO-EXPLORER solves 11 only.
Finally, we discuss that the approach is generic in the sense that RZS is applicable to solving many other goal-achieving problems for board games. 
\end{abstract}

\section{Introduction}\label{sec:introduction}

Traditional board games such as Go and Hex have played an important role in the development of artificial intelligence. 
Goal-achieving problems are puzzles that challenge players to achieve specific goals under given board configurations.

\emph{Life-and-death} (L\&D) problems in Go is a typical goal-achieving problem.
In Go, where the goal for both players is to hold more territory on the game board than their opponent, a critical skill is to identify safety for stone groups. 
A safe (live) stone group holds territory on the board indefinitely, thereby giving an advantage to the player that owns it.
To build this skill, Go players have been creating and solving L\&D problems, also called {\em tsumego}, for centuries.
Many of these problems are challenging even for professional players. 
A straightforward method of solving L\&D problems using computer programs involves finding a solution tree \cite{stockman1979minimax,pijls2001game} that represents a full strategy of answering moves. 
As a computational problem, one of the main challenges is that the large branching factor in Go tends to lead to a prohibitively large solution tree. 
To cut down on the search size, Kishimoto and M{\"u}ller \shortcite{kishimoto2003df,kishimoto2005dynamic,kishimoto2005search} manually designated a specific search space to prevent searching irrelevant spaces.

An elegant way to reduce the branching factor is to take advantage of threats \cite{allis1994searching}, which are moves that force the opponent to respond in a specific way.
When threats are involved, not all opponent moves need to be explored to ensure correctness when solving L\&D problems, especially moves that do not answer the threat.
In this vein, Thomsen \shortcite{thomsen2000lambda} proposed a threat-based search algorithm named {\em lambda search}, by trying null moves first (usually by passing), then deriving a so-called {\em relevance zone (RZ)}, outside of which the opponent does not need to search.

This paper proposes a novel approach called {\em RZ-Based Search (RZS)} to solving L\&D problems.
We illustrate the basic concepts of RZS with examples in Go and Hex. 
RZS itself is then described in detail with Go examples.
With RZS, moves are searched first, before they are determined to be null moves post-hoc.
This allows our approach to be seamlessly incorporated into most search algorithms.
Specifically for this paper, we applied RZS into an AlphaZero-like program \cite{silver2018general}, well-known for playing at super-human levels.

A straightforward way of leveraging AlphaZero methods for goal-achieving problems is to use it in the construction of solution trees, by adding Boolean AND-OR tree logic for wins and losses into the AlphaZero MCTS procedure.
However, AlphaZero is fundamentally trained for the task of winning the game, rather than proving it exhaustively. 
This leads to a discrepancy in behaviour between a strong player and a solver. 
Namely, AlphaZero picks any winning move, rather than the move that leads to the shortest distance to win, as also mentioned in \cite{agostinelli2019solving}. 

In order to reduce the solution tree size (with shorter paths to prove), we also propose a new AlphaZero-like training method called {\em Faster to Life (FTL)}, by modifying the goal of the AlphaZero algorithm so that it prefers winning in fewer moves.
In our experiments, among a collection of 20 7x7 L\&D problems, all are solved with RZS integrated into AlphaZero with FTL; with RZS without FTL, 5 can be solved; without RZS nor FTL, none can be solved.
Furthermore, among 106 19x19 Go problems selected from a well-known L\&D book, written by a Go master, we solve 68 with both FTL and RZS, while solving 36 with RZS but without FTL.
In contrast, a previous solver by \citet{kishimoto2005search} that does not use FTL and RZS solves 11 problems. 

\section{Background}\label{sec:background}

\subsection{Solution Trees for Achieving Goals}
\label{subsec:solution_trees}
For two-player games, a solution tree \cite{stockman1979minimax,pijls2001game} is a kind of AND-OR search tree representing a full strategy of answering moves to achieve a given goal, e.g., safety of pieces for Go or connectivity of pieces for Hex, as described in next subsections. 
Both AND and OR nodes represent game positions, and edges represent moves from one position to another. 
In this paper, for a position $p$, let $\beta(p)$ denote the corresponding board configuration and $\pi(p)$ the player to play. 
The player to play at an OR node is called the OR-player, denoted by \ORplayer, whose objective is to achieve the given goal, while the other player, called the AND-player, denoted by \ANDplayer, tries to prevent the same goal from happening.
Achieving the goal in this context is equivalent to {\em winning} from the OR-player's perspective.

A solution tree, rooted at $n$, is a part of a search tree that satisfies the following properties. 
\begin{enumerate}
\item If $n$ is a leaf, the goal is achieved for \ORplayer at $n$.

\item If $n$ is an internal OR-node, there exists at least one child whose subtree is a solution tree. For simplicity, we usually consider the case where there is exactly one child. 

\item If $n$ is an internal AND-node, all legal moves from $n$ must be included in the solution tree, and all subtrees rooted from $n$'s children are solution trees. 
\end{enumerate}

\begin{figure}[th]
\centering
\begin{subfigure}[t]{0.24\columnwidth}\includegraphics[width=\columnwidth]{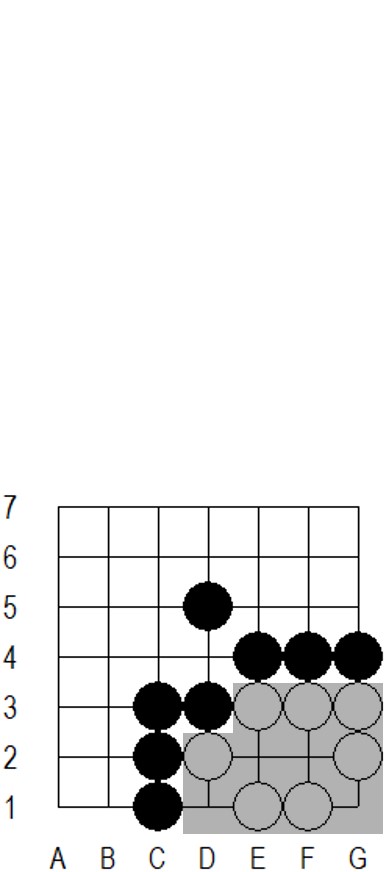}\caption{$p_a$}\end{subfigure}
\begin{subfigure}[t]{0.24\columnwidth}\includegraphics[width=\columnwidth]{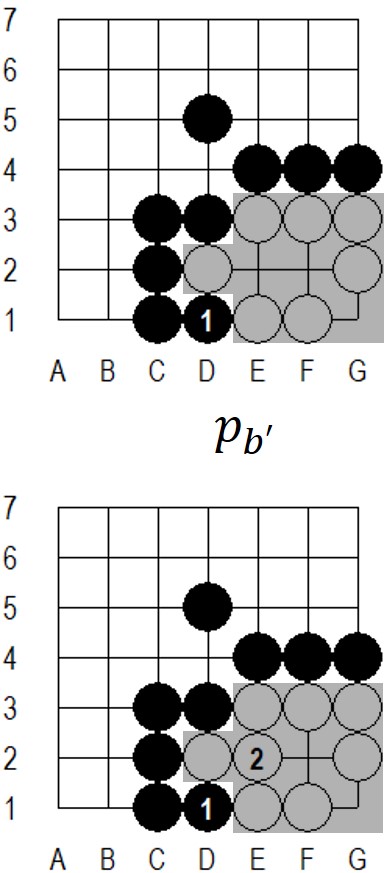}\caption{
$p_b$}\end{subfigure}
\begin{subfigure}[t]{0.24\columnwidth}\includegraphics[width=\columnwidth]{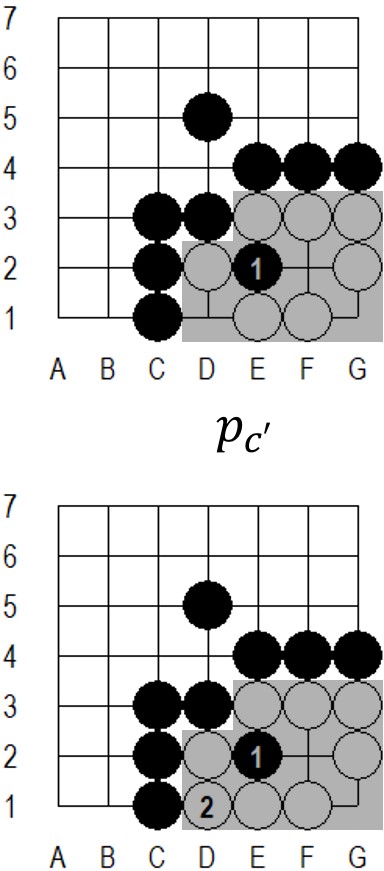}\caption{
$p_c$}\end{subfigure}
\begin{subfigure}[t]{0.24\columnwidth}\includegraphics[width=\columnwidth]{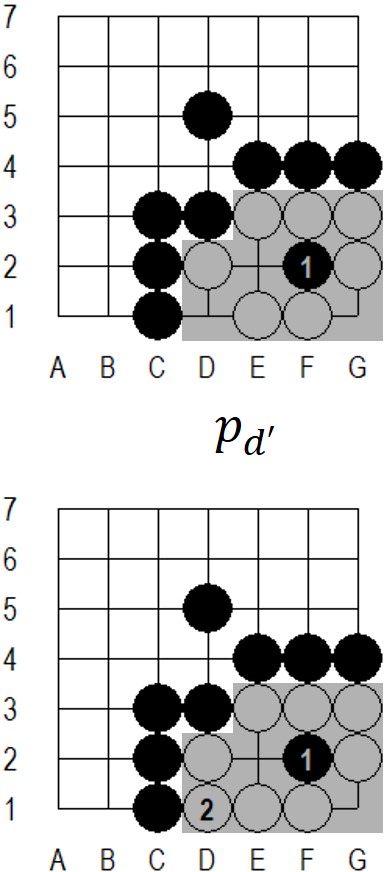}\caption{
$p_d$}\end{subfigure}    
\caption{
An example for Go. For position $p_a$, if Black plays at 1 as in $p_{b'}$, $p_{c'}$, and $p_{d'}$, White replies to win at 2 as in $p_b$, $p_c$, and $p_d$ respectively. 
Since positions $p_b$, $p_c$, and $p_d$ satisfy UCA, we obtain RZs as the shaded cells, denoted by $z_b$, $z_c$, and $z_d$ respectively. 
For their previous positions $p_{b'}$, $p_{c'}$, and $p_{d'}$, White wins by playing 2 and it is not hard to see that their corresponding RZs are also $z_b$, $z_c$, and $z_d$ respectively. 
Furthermore, consider $p_a$. For all moves outside $z_a$, denoting the shaded zone of $p_a$, White can simply play at E2 to win. 
Thus, $p_a$ wins. 
Note that G1 is illegal for Black since suicide is prohibited. 
In addition, $z_a$ is its RZ, a union of $z_b$, $z_c$, and $z_d$, since White follows the same strategy as above to win for all Black moves outside of it. 
}
\label{fig:benson_examples}
\end{figure}

\subsection{L\&D Problems for Go}
\label{subsec:go_term_uca}

We first quickly review the game of Go and then define the goal for L\&D problems. 
The game of Go is a game played by two players, Black and White, usually on a $n\times n$ square board, where at most one stone, either black or white, can be placed on each intersection of the board, called a {\em cell} in this paper.
A set of stones of the same color that are connected via adjacency to one another in four directions is called a {\em block}. 
Unoccupied cells adjacent to a block are called {\em liberties} of that block. 
A block is captured if an opponent stone is placed in the block's last liberty.
Following the rules of Go, each block has at least one liberty, and players are not allowed to make suicidal moves that deprives the last liberty of their own blocks, unless that move also captures opponent stones.
{\em Safety} refers to a situation where blocks cannot be captured by the opponent under optimal play. 

A L\&D problem is defined as follows. 
Given a position and some stones of \ORplayer designated as crucial, as \cite{kishimoto2005search}, \ORplayer is said to achieve the goal if safety can be guaranteed for any of these crucial stones, or fail if all crucial stones are captured. 

A set of blocks is said to be {\em unconditionally alive (UCA)} by \citet{benson1976life} if the opponent cannot capture it even when unlimited consecutive moves are allowed.
Namely, each of these blocks can sustain at least two liberties for unlimited consecutive moves by the opponent. 
By illustrating the seven positions in Figure \ref{fig:benson_examples}, the white blocks satisfy UCA in $p_b$, $p_c$, and $p_d$, illustrated in the bottom row of (b), (c), and (d), but not for others.
The details of UCA are specified by \citet{benson1976life}.
Achieving UCA for any crucial stones is a valid solution to a L\&D problem.

In addition to UCA, safety can be achieved through other ways, such as seki \cite{niu2005recognizing}, ko, or situational super-ko (SSK) \cite{van2003solving}. 
Coexistence in seki is also considered a win for \ORplayer, since the crucial stones are safe from being captured. 
Lastly, the rules of ko or SSK are used to prohibit position repetition, i.e., any move that repeats a previous board position with the same player to play is illegal.
In this paper, we limit our goal to achieving safety via UCA only for simplicity. 
L\&D problems in which the only solution is via seki and ko (SSK) are left as open problems.
For simplicity, achieving safety via UCA is referred to as a {\em win}, or {\em winning} for the rest of this paper.

A specific instance of the more general L\&D definition above is the so-called 7x7 kill-all Go game, in which Black, the first player, plays two stones initially, then both players play one stone per turn alternately on a 7x7 Go board. 
The goal of White, playing as \ORplayer, is to achieve safety for any set of white stones via UCA, while Black (playing as \ANDplayer) must kill all white stones.
Note that for this specific problem, there is no need to mark crucial stones for White; any safe white stone is sufficient.
For simplicity, the rest of this paper is written with 7x7 kill-all Go in mind, i.e., \ORplayer wins when any white stones are UCA, unless specified otherwise. 

\subsection{Connectivity for Hex}

For generality, we also discuss another goal-achieving problem for Hex. 
We first review Hex, then define the goal of achieving connectivity. 
Hex is a two-player game commonly played on a board with $n\times n$ hexagonal cells in a parallelogram shape as illustrated in Figure \ref{fig:hex_examples} below.
The two players, Black and White, are each assigned two opposite sides of the four boundaries of the board, and take turns to place stones of their own color. 
Similarly, a set of stones of the same color that are connected via adjacency to one another in a hexagonal way is called a {\em block}, and each side of the board can be viewed as a block of the same color for simplicity.
A board configuration consists of four sides and all cells, each of which is unoccupied or occupied by either Black or White. 
The player who has a block of their own color connecting the two correspondingly colored sides wins the game. 
The rules of Hex are relatively simple in the sense that there are no stone capturing (i.e., any changes to stones that are already on the board) and no prohibited moves like suicides.

A goal-achieving problem for Hex is defined as follows. 
Given a position $p$ and some stones of \ORplayer designated as crucial, \ORplayer is said to achieve the goal if they can connect these crucial stones together, or fail otherwise.
For example, in Figure \ref{fig:hex_examples} (a), a problem may specify that with Black (\ANDplayer) playing first, C3 is a crucial stone and White must connect it to the lower side of the board. 
Figure \ref{fig:hex_examples} (c) shows a case of achieving the goal. 

\section{Our Approach}\label{sec:approach}

This section first defines relevance zones (RZs), and then reviews null moves and must-play regions \cite{hayward2003solving,hayward2006hex,hayward2009puzzling} with illustrations.
Next, we define {\em RZ-based solution trees (RZSTs)} and present {\em Consistent Replay Conditions}, under which \ORplayer can replay and therefore reuse previously searched solution trees to win. 
The final subsection proposes a novel approach, called {\em RZ-based Search (RZS)}, which can be incorporated with other depth-first or best-first search for solving goal-achieving problems.

\subsection{Relevance Zones}
\label{subsec:rzone}

A zone $z$ is a set of cells on the board. 
Given a position $p$ and a zone $z$, let $\beta(p)\odot z$ denote the board configuration $\beta(p)$ inside the zone $z$, also called a {\em zone pattern}.
Let $\bar{z}$ denote the zone outside of $z$. 
Thus, $\beta(p)\odot \bar{z}$ is the zone pattern outside the mask $z$. 

A zone $z$ is called a {\em relevance-zone} ({\em RZ}) with respect to a winning position $p$, if the following property is satisfied.
\begin{description}
\item[RZ-1] 
For all positions $p_*$ with $\beta(p_*) \odot z = \beta(p)\odot z$ (the same zone pattern) and $\pi(p_*)=\pi(p)$ (the same player turn), $p_*$ are also wins.
\end{description}

\begin{figure}[t]
    \centering
    \begin{subfigure}[t]{0.57\columnwidth}\includegraphics[width=\columnwidth]{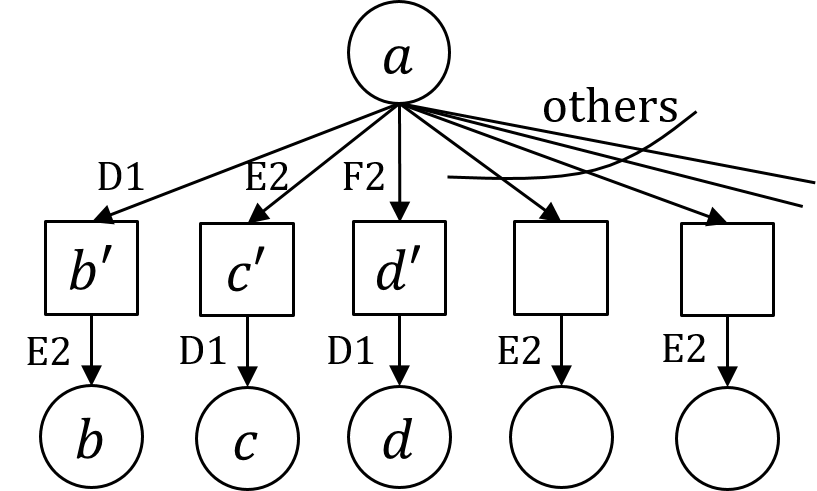}\caption{The solution tree of Figure \ref{fig:benson_examples} (a).}\end{subfigure}
    \begin{subfigure}[t]{0.42\columnwidth}\includegraphics[width=\columnwidth]{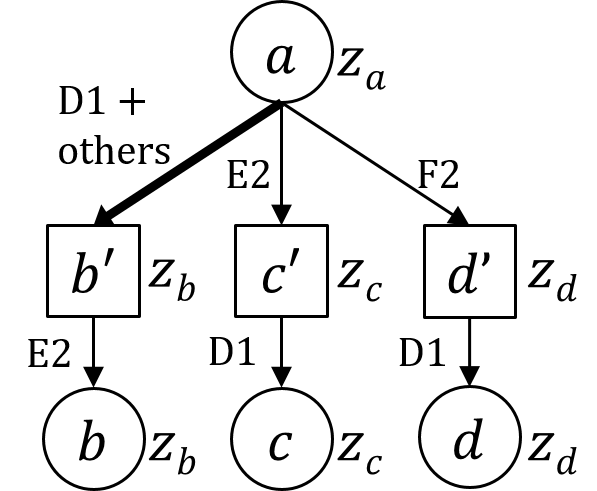}\caption{The RZST for (a).}\end{subfigure}
    \caption{Solution trees.}
    \label{fig:sol_trees}
\end{figure}

In general, there exists some RZ with respect to a winning position. 
For positions that achieve UCA for Go, their RZs include all cells of the safe blocks and their regions \cite{benson1976life}, since these blocks are safe regardless of any moves or changes outside the zone.
Examples are given in Figure \ref{fig:benson_examples}.
Thus, a solution tree for $p_a$ is shown in Figure \ref{fig:sol_trees} (a), and the zone $z_a$ is clearly an RZ for $p_a$. 
From RZ-1, we can easily obtain the following property. 
\begin{description}
\item[ZX-1] If $z$ is an RZ with respect to $p$, any bigger zone $z' \supseteq z$ is an RZ for $p$ as well.
\end{description}
For example, the RZ for $p_a$ in Figure \ref{fig:benson_examples} (a) can be expanded, say, by adding the black blocks or any number of unoccupied cells.
Taken to the extreme, as long as $p$ is a win, the entire board can serve as an RZ.

\subsection{Null Moves and Must-Play Regions}
\label{subsec:must_play}


Relevance zones can be derived based on null moves, as in lambda search \cite{thomsen2000lambda,wu2010relevance}, and must-play regions \cite{hayward2003solving,hayward2006hex,hayward2009puzzling}. 
For example, in Figure \ref{fig:benson_examples}, if Black makes a null move on $p_a$ (in this case any move outside of $z_a$, including passes), White simply plays at E2 to reach UCA, obtaining in the process an RZ, $z_b$, like $p_b$. 
That is, it is a win for White regardless of any moves or changes outside $z_b$.
Thus, back to $p_a$, to prevent White from winning directly, Black {\em must play} (or {\em try}) at unoccupied cells inside $z_b$, including moves E2, F2 and G1; $z_b$ is conceptually equivalent to {\em must-play regions} in Hex by \citet{hayward2003solving}. 
By ignoring moves outside $z_b$, the search space is greatly reduced; we refer to this branch reduction as {\em relevance-zone pruning}. 
Since G1 is illegal in Go, only E2 and F2 actually need to be tried.
For Hex, all the unoccupied cells inside an RZ are legal and are required to be tried.

For RZ pruning, previous lambda-based search methods tried null moves first, then derived RZs as above \cite{thomsen2000lambda,yoshizoe2007lambda,wu2010relevance}. 
However, the decision on when and how to play null moves is itself a heuristic that requires deliberate thought. Usually, a null move consists of passing with the hope of reducing the branching factor. Since giving up one's turn tends to be one of the weakest moves to consider in most games, this runs counter to the general intuition that strong moves should be searched ﬁrst.
In addition, when the null move does not result in pruning, the extra effort spent to search the null move may be entirely wasted.
Note that many previous researches employed some lightweight heuristics \cite{kishimoto2003df,yoshizoe2007lambda} for forced moves to prevent from incurring too much overhead. 

In this paper, we propose searching promising moves directly without needing to try null moves first, then determining which moves are null moves post-hoc. 
For example, in Figure \ref{fig:benson_examples} (a), let Black attempt to kill White's blocks by playing at D1 as in (b), leading to White responding at E2 with UCA achieved and the RZ $z_b$, which encompasses the cells with UCA white stones. 
If we now look back at Black's decision of playing at D1, it is outside the RZ $z_b$, which makes it in effect a null move.
According to Property RZ-1, White wins regardless of any moves and changes (including the move D1) outside $z_b$. 

\begin{figure}[t]
\centering
\begin{subfigure}[t]{0.25\columnwidth}\includegraphics[width=\columnwidth]{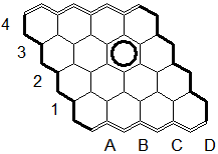}\caption{}\end{subfigure}
\begin{subfigure}[t]{0.25\columnwidth}\includegraphics[width=\columnwidth]{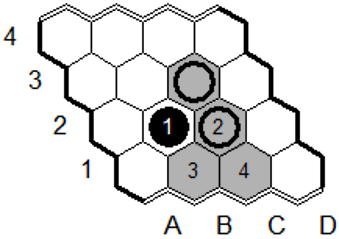}\caption{}\end{subfigure}    
\begin{subfigure}[t]{0.25\columnwidth}\includegraphics[width=\columnwidth]{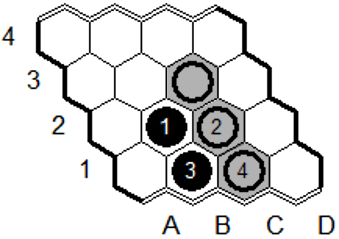}\caption{}\end{subfigure}
\\
\begin{subfigure}[t]{0.25\columnwidth}\includegraphics[width=\columnwidth]{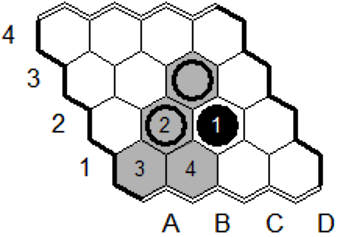}\caption{}\end{subfigure} 
\begin{subfigure}[t]{0.25\columnwidth}\includegraphics[width=\columnwidth]{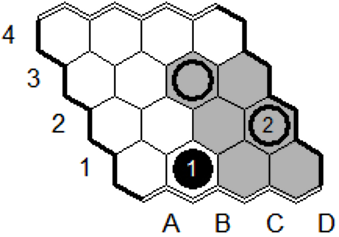}\caption{}\end{subfigure}
\begin{subfigure}[t]{0.25\columnwidth}\includegraphics[width=\columnwidth]{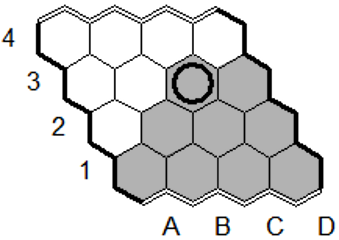}\caption{}\end{subfigure}
\caption{
An illustration of achieving connection between C3 to the bottom side in Hex. 
For Black B2 in (b), White replies at C2, then wins by either B1 or C1.
The RZ is the shaded cells in (b), since the goal can be achieved regardless of any moves outside of the RZ. 
Similar to the example in Figure \ref{fig:benson_examples}, B2 is viewed as a null move, and the must-play region for Black is reduced to include only the unoccupied cells, C2, B1, and C1.
Next, if Black plays at C2 as in (d), White replies at B2 and wins due to symmetry to (b). 
Again, since Black C2 is a null move, the RZ is the shaded cells as in (d) and the must-play region is reduced to B1 only, from the intersection of the two RZs in (b) and (d).
Now, we only need to search Black B1, in which case White replies at D2 for a win as in (e), since C3 is guaranteed to be connected to D2, and D2 is guaranteed to be connected to the lower side independently.
Since all unoccupied cells in the must-play region have been searched, the goal can be achieved, and the RZ for the position in (a) is the shaded area in (f), which is the union of the above RZs, denoted by $z_f$.
Namely, C3 can be connected to the lower side, regardless of any moves or changes outside of $z_f$. 
}
\label{fig:hex_examples}
\end{figure}

In this case, the must-play region that initially includes all unoccupied cells is reduced to E2 and F2 (note that G1 is illegal). 
If Black plays at E2 as in (c), it is a win with RZ $z_c$ for White's reply at D1. 
Since E2 is inside $z_c$, it is not a null move and only E2 can be removed from the must-play region. 
It is similar for Black at F2 in (d) with an RZ $z_d$. 
After searching E2 and F2, the must-play region becomes empty, which implies a win for White for the position $p_a$ in (a). 
The RZ $z_a$ for $p_a$ is therefore the union of $z_b$, $z_c$, and $z_d$. 
The power of relevance zones and must-play regions are illustrated in more examples for Hex and Go as follows. 

\begin{figure}[t!h!]
\centering
\begin{subfigure}[t]{0.24\columnwidth}\includegraphics[width=\columnwidth]{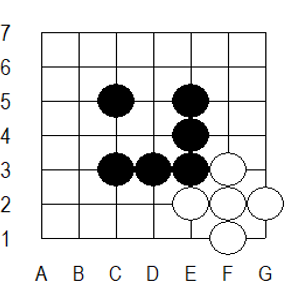}\caption{}\end{subfigure}
\begin{subfigure}[t]{0.24\columnwidth}\includegraphics[width=\columnwidth]{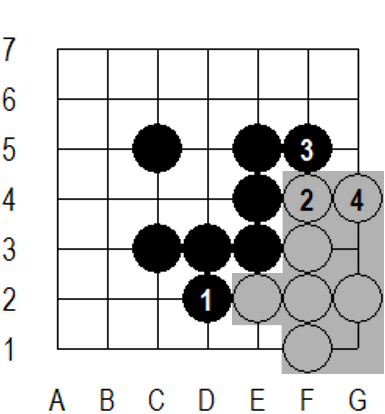}\caption{}\end{subfigure}
\begin{subfigure}[t]{0.24\columnwidth}\includegraphics[width=\columnwidth]{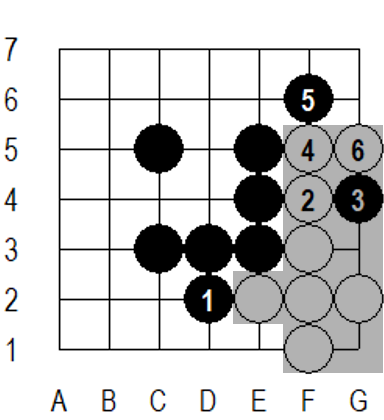}\caption{}\end{subfigure}
\begin{subfigure}[t]{0.24\columnwidth}\includegraphics[width=\columnwidth]{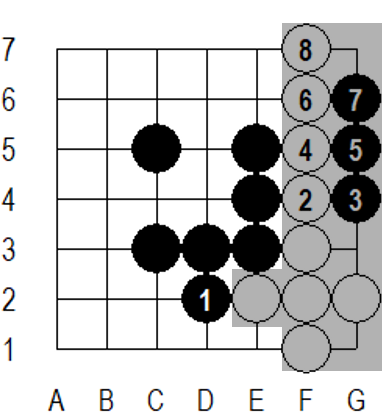}\caption{}\end{subfigure}
\begin{subfigure}[t]{0.24\columnwidth}\includegraphics[width=\columnwidth]{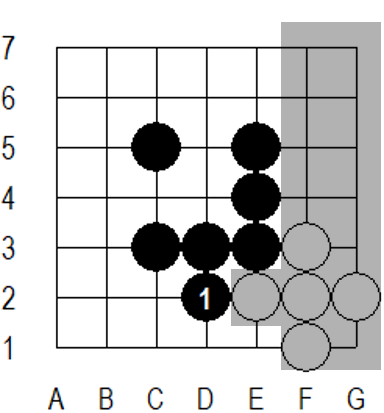}\caption{}\end{subfigure}
\begin{subfigure}[t]{0.24\columnwidth}\includegraphics[width=\columnwidth]{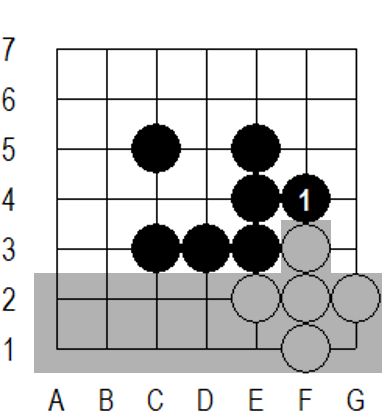}\caption{}\end{subfigure}
\begin{subfigure}[t]{0.24\columnwidth}\includegraphics[width=\columnwidth]{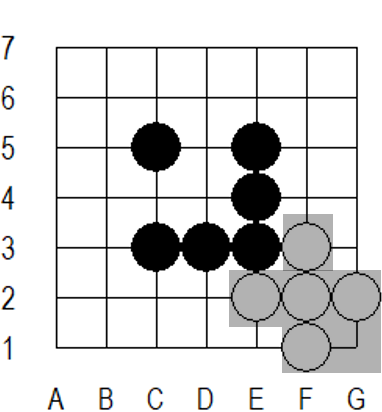}\caption{}\end{subfigure}  
\begin{subfigure}[t]{0.24\columnwidth}\includegraphics[width=\columnwidth]{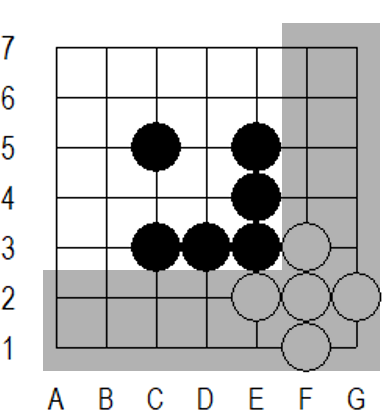}\caption{}\end{subfigure}
\caption{A 7x7 kill-all Go example. 
First, Black tries to prevent White from achieving UCA by playing at, say, D2 as in (b). 
In this case, White replies at F4 in order to achieve UCA by making a second region at G3 (the first is at G1). 
If Black counters with F5, then White achieves UCA with G4, where the set of shaded cells is its RZ, denoted by $z_b$. 
Since F5 is outside of $z_b$, it is a null-move, so the must-play region includes the two legal moves G3 and G4 that Black can try. 
For G3, it is trivial to win at G4. 
Black now considers playing at G4 instead, leading to the position in (c). 
In this variation, White plays at F5 in response to Black's move at G4. 
Again, if Black counters with F6, then White achieves UCA at G5.
This search continues with Black being forced to consider G5, and again White at F6 forces Black to play at G6 as in (d).
Finally, White plays at F7 to achieve UCA. 
Thus, for the Black move at D2, we can conclude that White wins and the RZ $z_e$ is the union of the RZs as in (e).
Interestingly, since D2 is outside of $z_e$, it is a null move and all of Black's moves outside $z_e$ can be disregarded. 
The must-play region is therefore the unoccupied cells in $z_e$.
Now, let Black choose to play F4 (inside $z_e$) as in (f). Similarly, White wins with an RZ $z_f$ as in (f) due to symmetry to (e). 
Since we can disregard Black moves outside $z_e$ and $z_f$, the must-play region becomes the intersection of the two RZs as shown in (g), which contains no legal moves for Black. 
Thus we have proved White wins in (a), with large reductions to the search space.
Note that the corresponding RZ for the position in (a) is the union of the two RZs, shaded in (h).}
\label{fig:cross_example}
\end{figure}

\subsubsection{Illustrations}
We first illustrate an example for the game of Hex to facilitate to understand. 
Previously, \cite{hayward2003solving} has used the above notion of null moves and must-play regions to solve Hex problems.
An example for Hex is given in Figure \ref{fig:hex_examples} (a), modified from \cite{hayward2003solving}, where its goal is to connect C3 to the lower side, and can be achieved through Figure \ref{fig:hex_examples} (b)-(f). 

 Now, we want to illustrate another example to demonstrate the power of RZ pruning for the game of Go using the above approach. 
The position in Figure \ref{fig:cross_example} (a) can be derived as a win for White as shown in (b)-(h). 
Thus, we can prove White wins in (a) by only searching two moves at D2 and F4 as shown in (e) and (f) respectively.

\subsubsection{Comparison with Previous Null Move Methods}
Other than Hex, most traditional RZ approaches need to search null moves in advance to facilitate pruning, e.g. null-move pruning in chess \cite{donninger1993null,heinz1999adaptive,campbell2002deep} and lambda search \cite{thomsen2000lambda,yoshizoe2007lambda,wu2010relevance}. 
For example, for the position in Figure \ref{fig:cross_example} (a), if we want to make another region, say at G3, two consecutive null moves for Black need to be searched first (say, two passes) for White to play consecutively at F4 and G4, as in second order lambda search \cite{thomsen2000lambda}. 
Null moves tend to be weak choices for \ANDplayer (Black in this case), which runs counter to the intuition that strong moves should be searched first.
Consequently, heuristics are usually used to determine whether and how \ANDplayer makes null moves. 
During this process, incorrect guesses for null moves may incur additional overhead. 

In contrast, with our approach, \ANDplayer searches ahead, then retroactively handles null moves after solving them. 
In most cases of winning, \ANDplayer gradually search those moves far away from the fighting regions, that is, it is likely to play null moves which are located outside of RZs. 
This allows our method to be seamlessly and elegantly incorporated into search algorithms such as MCTS \cite{coulom2006efficient} or alpha-beta search \cite{knuth1975analysis}; namely, the algorithm can prioritize strong moves as usual, and exploit RZ pruning simultaneously.

\subsection{Relevance-Zone Based Solution Trees (RZSTs)}
\label{subsec:pure-rzst}

The previous subsection illustrates the power of using relevance zones.
We now incorporate the notion of relevance zones into solution trees recursively, which we refer to as Relevance-Zone Based Solution Trees (RZSTs).
Since Hex is relatively straight-forward compared to Go, we will focus on Go in the rest of this section. 

For a node $n$, let $p(n)$ be its corresponding position of $n$. 
For simplicity, let $\beta(p(n))$ be simplified as $\beta(n)$, and $\pi(p(n))$ as $\pi(n)$.
If $p(n)$ is a win, let $z$ be an RZ for $n$, where $z$ is guaranteed to exist as described above.
From the definition of RZs, since the conditions of the win are satisfied entirely by the zone pattern within $z$, and the pattern in $\bar{z}$ are irrelevant, the solution tree rooted at $n$ does not need to include any children in $\bar{z}$ if $n$ is an AND-node. 
As an example, the solution tree in Figure \ref{fig:sol_trees} (a) can be reduced to include the first three children only, as shown in (b), where the thicker arrow for the move D1 indicates a null move whose RZ $z_b$ is used to prune all moves other than E2 and F2. 
More specifically, all the moves outside $z_b$ lead node $a$ to $b'$.

The above definition for RZSTs leaves out how RZs are derived for the winning node $n$.
A key to justifying the correctness of RZs is to replay RZSTs rooted at null moves. 
For the example in Figure \ref{fig:benson_examples} (a), if Black plays anywhere outside $z_a$, White can win by simply "replaying" the strategy depicted by the solution tree at $b'$, which is simply E2.
Similarly, for another example in Figure \ref{fig:cross_example} (e): if Black plays anywhere outside the shaded RZ (say C2), White can win by simply "replaying" F4, as shown in (b)-(d).

Before discussing this in more detail in later subsections, we first describe a common principle of deriving RZs.
\begin{description}
\item[ZZ-1] For parent-child pairs $n$ and $n'$ in RZSTs, let $z$ and $z'$ be their RZs respectively. 
We derive both RZs such that $z'\subseteq z$. 
\end{description}
Intuitively, since the cells in $\bar{z}$ are irrelevant for $n$, moves in $\bar{z}$ should also be irrelevant to $n$'s descendants.
We can see this from many examples in the previous subsection, where the RZ is a union of their children's RZs.

\subsection{Consistent-Replay (CR) Conditions}
\label{subsec:crproperty}

This subsection presents some {\em sufficient conditions} for RZs such that replay is always legal, which in turn justifies the correctness for RZSTs.
An important notion of RZs, which originates from threat-space search \cite{thomsen2000lambda,wu2010relevance}, is that \ANDplayer moves played outside the zone do not interfere with the winning strategy of \ORplayer inside the zone, which is represented by RZSTs.

Let a winning node $n$ be the root of an RZST, and $z$ be an RZ for $n$.
From Property RZ-1, all positions $p_*$ with $\beta(p_*) \odot z = \beta(n) \odot z$ and $\pi(p_*)=\pi(n)$ are wins too. 
In addition to UCA, one way to justify the correctness for $p_*$ is to verify that the winning strategy of $p_*$ consists of simply following (or replaying) the RZST to win.
In order to replay consistently for all $p_*$ based on the RZST, we propose the following sufficient conditions, called {\em Consistent Replay (CR) Conditions}.

\begin{description}
\item[CR-1] Suppose it is \ANDplayer's turn.
Consider all \ANDplayer's legal moves at unoccupied cells $m$ inside $z$ on $p_*$. 
Then, playing at $m$ on $p(n)$ must be legal for \ANDplayer as well, and must lead to the same zone pattern. Namely, $\beta(p')\odot z=\beta(p'_*) \odot z$, where $p'$ and $p'_{*}$ are the next positions of $p(n)$ and $p_*$ respectively after the move at $m$. 

\item[CR-2] Suppose it is \ANDplayer's turn.
For any \ANDplayer's move outside $z$ on $p_*$, the zone pattern inside $z$ remains unchanged, namely, $\beta(p) \odot z = \beta(p_*) \odot z = \beta(p'_*)\odot z$, where $p'_{*}$ is the next position of $p_*$ after the move. 

\item[CR-3] Suppose it is \ORplayer's turn and the winning move is at an unoccupied cell $m$ inside $z$ on $p(n)$. Then, the move at $m$ must be legal on $p_*$ for \ORplayer as well, and must lead to the same zone pattern. Namely, like CR-1, $\beta(p')\odot z=\beta(p'_*) \odot z$, where $p'$ and $p'_{*}$ are the next positions of $p(n)$ and $p_*$ respectively after the move at $m$. 

\end{description}

The conditions CR-1 and CR-3 ensure that \ORplayer can replay the same winning responses to all \ANDplayer moves inside $z$ on $p_*$, while maintaining identical RZ patterns.
This can be illustrated by the winning position $p_a$ with the RZ $z_a$ in Figure \ref{fig:benson_examples} (a). 
For all positions $p_*$ with the same RZ pattern as $p_a$, all \ANDplayer legal moves inside the zone on $p_*$ (e.g., D1, E2, and F2) are also legal on $p_a$, so \ORplayer can play winning responses in $p_*$ as those in Figure \ref{fig:benson_examples} (b), (c) and (d), and the resulting zone patterns remains the same.

CR-2 ensures that for \ANDplayer moves outside $z$, the zone pattern inside $z$ remains unchanged. 
In this case, \ORplayer simply follows the same winning strategy as that for $n$. 
From ZZ-1, there must exist some null move with a zone $z' \subseteq z$ in $n$, so White can simply respond by replaying the same strategy as that for $n'$. 
For example, in Figure \ref{fig:benson_examples} (a), for all moves outside the RZ $z_a$, \ORplayer simply chooses a null move, D1 in this case, and then uses its reply at E2 to reply in $p_*$. 
From the above, we obtain the following lemma. 

\begin{lemma}
Assume that node $n$ is a win with an RZ $z$, and that all RZs in the RZST rooted at $n$ are derived following the three CR conditions. 
Then, for all positions $p_*$ with $\beta(p_*) \odot z = \beta(n)\odot z$ (the same zone pattern) and $\pi(p_*)=\pi(n)$ (the same player turn), $p_*$ is a win by following the winning strategy of the RZST.
\label{lemma:consistent-replay}
\end{lemma}

\begin{proof}
It suffices to show that \ORplayer can replay the winning strategy of the RZST rooted at $n$ by induction. 
If the goal is reached for $p(n)$ with an RZ $z$, then the goal is reached for $p_*$ too.

Now, assume that the goal has not been reached for $p(n)$ yet.
Consider the case of \ORplayer to play. 
Let \ORplayer win by playing a winning move $m$, leading to a new node $n'$.
From CR-3, \ORplayer can play the same winning move at $m$ on $p_*$ and to a new position $p'_*$. 
Also from CR-3, the resulting zone patterns are identical, i.e., $\beta(p'_*)\odot z=\beta(n') \odot z$.
Since $n'$ is also a win, there exists an RZ $z'$ for it, where $z' \subseteq z$ from ZZ-1. 
This implies that $\beta(n')\odot z'=\beta(p'_{*}) \odot z'$. 
From RZ-1, it is a win for $p'_*$ too. 

For the case of \ANDplayer to play, we need to consider all legal moves $m$ on $p_*$ by \ANDplayer. 
First, assume $m$ to be inside $z$. 
From CR-1, \ANDplayer can play at the same $m$ on $p(n)$, resulting in the same zone pattern, i.e., $\beta(n')\odot z=\beta(p'_{*}) \odot z$ where $n'$ is the next node of $n$ and $p'_*$ is the next position of $p_*$.
Since $n'$ is also a win, there exists an RZ $z'$, with $z' \subseteq z$ from ZZ-1, such that $\beta(n')\odot z'=\beta(p'_*) \odot z'$. 
Thus, it is a win for $p'_*$ by following the RZST rooted at $n'$. 

Second, assume $m$ to be outside of $z$. 
From CR-2, the zone pattern inside $z$ remains unchanged, namely, $\beta(n) \odot z = \beta(p_*) \odot z = \beta(p'_*)\odot z$, where $p'_*$ is the next position of $p_*$ after the move.
\ORplayer can win by simply following the RZST rooted at $n$. 
\end{proof}

In order to satisfy the above CR conditions, we first derive RZs from leaf nodes (e.g. by including all cells that contain the UCA blocks). For internal nodes, we dilate (expand) children RZs based on ZX-1 such that all CR conditions hold. 
In the worst case $z$ is dilated to include the whole board, in which case $\bar{z}$ is empty. 
Note that smaller RZs lead to smaller trees, so for a winning position, it is preferable to choose as small an RZ as possible for more RZ pruning. 
For the game of Go, dilation requires some thought for capturing stones and prohibiting suicidal moves. A heuristic \emph{zone dilation} method for Go, referred to as the function $dilate$($z$) below, is presented in \cite{github_url}.
In contrast, the game of Hex needs no dilation, since there are no stone capturing (or moving) and no prohibited moves like suicides.

\subsection{Relevance Zone Based Search}
\label{subsec:RZST}

In this subsection, we propose a goal-achieving search approach, called {\em RZ-based Search ({\em RZS})}, which can be embedded into other search algorithms using best-first search or depth-first search. 
In this approach, we propose a method to derive more efficient RZSTs, called {\em replayable RZSTs}, which are defined recursively. 
Namely, replayable RZSTs rooted at $n$ as well as their RZs $z$ will be constructed and derived, starting from leaf nodes, in a bottom-up manner.
The three CR conditions need to be satisfied if the node $n$ is a win with an RZ $z$. 
Our approach RZS can be incorporated into best-first search, such as MCTS, as described in the rest of this subsection, which is used in our experiments later. 
Actually, RZS can also be incorporated into depth-first search (e.g. alpha-beta search) as presented in \cite{github_url}, which behaves similarly to the must-play-based Hex solver by \citet{hayward2009puzzling}. 

\subsubsection*{Leaf Nodes}
Suppose that position $p(n)$ for generated node $n$ achieves the goal, e.g., UCA, and wins. 
Then, $n$ is an RZST (consisting of a single leaf node), and their RZ is derived accordingly. 

\subsubsection*{Internal Nodes}
Suppose in $p(n)$, no \ORplayer blocks are UCA yet (i.e., the position is not immediately winning). These internal nodes $n$ will be recursively updated from their children $n'$ (in a bottom-up manner), e.g., during the backpropagation phase of MCTS. 
Since $n$ is an internal node, it can either be an OR-node or an AND-node, described as follows.

\subsubsection*{{\em Internal OR-Nodes}}
Suppose it is \ORplayer's turn ($n$ is an OR-node).
Assume that the child $n'$ via a legal move $m$\footnote{In this context, $m$ is both a conceptual construct representing an edge in the tree, and also a cell point signifying where the player places their stone. In this way we may add $m$ into some zone $z$, which is itself a set of cells.} is proven to be a win, and that the subtree rooted at $n'$ is a replayable RZST.
Let $z'$ be the associated RZ for $n'$.
The RZ $z$ associated with $n$ is derived from $dilate(z \cup \{m\})$.
Thus, $n$ together with the RZST rooted at $n'$ forms a replayable RZST. 
From Lemma \ref{lemma:consistent-replay}, for all positions $p_*$ with $\beta(p_*)\odot z=\beta(n)\odot z$, \ORplayer is allowed to replay $m$ with the RZST rooted at $n'$ to win $p_*$ with the same next RZ $z'$.

\subsubsection*{{\em Internal AND-Nodes}}

Suppose it is \ANDplayer's turn.
Let $n$ maintain a must-play region $M$ that is initialized to all the legal moves in $p(n)$. 
Assume that a child $n'$ via a legal move $m \in M$ is proven as a win with the associated RZ $z'$, and that the subtree rooted at $n'$ is a replayable RZST. 
The node $n$ is updated as follows. 
\begin{enumerate}
    \item Consider the case that $m\notin z'$ and the move $m$ does not change the zone pattern inside $z'$, that is, $m$ is a null move. 
    Then, shrink the must-play region by $M=M \cap z'$. 

    \item Consider the case where $m\in z'$, that is, $m$ is not a null move. 
    Then, simply remove $m$ from $M$. 
\end{enumerate}
When the must-play region becomes empty, the node $n$ is a win with an RZ $z$, derived by $dilate(z_{union})$, where $z_{union}$ is the union of all the children's RZs. 
Then, $n$, together with all of its child RZSTs, forms a replayable RZST. 
From Lemma \ref{lemma:consistent-replay}, for all positions $p_*$ with $\beta(p_*)\odot z=\beta(n)\odot z$, \ORplayer is allowed to replay with the same winning strategy based on these child RZSTs. 

\section{Faster to Life}
\label{sec:ftl}

This section proposes an AlphaZero-like method, named {\em Faster to Life (FTL)}, so that the search prefers choosing moves that win with the least number of moves.
Consider 7x7 kill-all Go, where White wins as long as any white stones are UCA.
When using unaltered AlphaZero, we can simply set a komi\footnote{In Go, komi is a number of compensation points for White since Black has the advantage of playing the first move.} of 48.

With FTL, we defined the winning condition to be White achieving UCA for any number of stones within $d$ moves from the current position. 
For example, with $d=20$, White wins only if it is able to reach UCA within 20 moves.
Given multiple values of $d$, the problem setting is similar to determining win/loss with multiple komi values, which can be handled with multi-labelled value networks \cite{wu2018multilabeled}.

Given a position $p$, the value network head outputs a set of additional $d$-win rates (the rates of winning within $d$ moves), where $d$ ranges from $1$ to a sufficiently large number (set to 30 in our experiments). 
Namely, for all positions $p$ during the self-play portion of AlphaZero training, if White lives by UCA at the $d$-th move from the current position, we count one more $d'$-win for all $d' \geq d$, but not for all $d'<d$. 
These $d'$-wins are used to update the set of $d$-win rates of $p$. 
Hence, White tends to win faster if their $d$-win rates are high for low values of $d$.

To ensure self-play plays reasonably well under multiple values of $d$, we apply a concept similar to {\em dynamic komi} for multiple win rates \cite{baudivs2011balancing,wu2018multilabeled}, where we set the winning condition to $d_w$-win, and $d_w$ is adjusted dynamically during self-play. 
Usually, $d_w$ is adjusted such that the win rate is close to 50\% for balancing \cite{baudivs2011balancing}.
The above can be applied to 19x19 L\&D problems, with the difference that a win is defined as achieving UCA for any crucial stones.  

\section{Experiments}\label{sec:experiments}

To demonstrate our RZ solver, we solve a collection of L\&D problems on 7x7 kill-all Go and 19x19 Go.

\subsection{7x7 Kill-All Go L\&D Problems}

First, based on our program CGI \cite{wu2020accelerating}, we train two AlphaZero programs for 7x7 kill-all Go, one with FTL and the other without FTL method, and chose 20 problems for 7x7 kill-all Go for analysis.
A problem is marked as proven if a program can prove it within 500,000 simulations.
In addition, a transposition table is used to record proven positions to prevent redundant search. 

\begin{table}[tbp]
    \centering
    \begin{tabular}{|c|c|r|r|r|}
        \hline
        \multicolumn{2}{|c|}{} & w/o RZS & w/ RZS \\
        \hline
        \multirow{2}{*}{7x7} & w/o FTL & 0/20 & 5/20 \\
        \cline{2-4}
        & w/ FTL & 3/20 & \textbf{20/20} \\
        \hline
        \multirow{3}{*}{19x19} & w/o FTL (ELF) & 0/106 & 36/106 \\
        \cline{2-4}
        & w/ FTL & 6/106 & \textbf{68/106} \\
        \cline{2-4}
        & T-EXP & 11/106 & - \\  
        \hline
    \end{tabular}
    \caption{Number of solved problems under different settings.}
    \label{tab:rzone_experiments}
\end{table}

In this experiment, we compare versions with and without FTL and RZS.
The results, shown in Table \ref{tab:rzone_experiments}, indicate that RZS together with FTL solves all 20 problems. 

\subsection{19x19 Go L\&D Problems}
Next, we apply the RZS method to 19x19 Go L\&D problems, which are widely-studied in the Go community.
We select problems from a well-known L\&D problems book, aptly named the {\em "Life and Death Dictionary"}, written by Cho Chikun [\citeyear{cho1987life}], a Go grandmaster; problem difficulties range from beginner to professional levels. 
A total of 106 problems, where the goal is to keep any crucial stones alive with UCA, were chosen. 
For simplicity, these problems do not require seki or ko (SSK) to achieve safety.

We first incorporated our approach into a 19x19 Go MCTS program that uses a pre-trained network with 20 residual blocks from the open-source program ELF OpenGo \cite{Tian2019ELFOA}.
Second, we trained a 19x19 AlphaZero Go program with the FTL method by using the problems from the tsumego book {\em The Training of Life and Death Problems in Go} \cite{shao1991training}. The game ends immediately if Black/White solves the L\&D problems.
We also added two additional feature planes to represent the crucial stones of each problem for both players.

We set up programs in a similar manner to those in 7x7 kill-all Go.
The baseline is the program TSUMEGO EXPLORER \cite{kishimoto2005search} (abbr. T-EXP), which is a DF-PN based program with Go specific knowledge, for which the search space needs to be manually designated. 
Since the problems cannot be used directly for ELF OpenGo (the network is trained to play with a komi of 7.5) and T-EXP (it requires limited regions surrounded by \ANDplayer's stones), we modified the problems such that they satisfy the requirements of these two programs.
The time budget for each move is limited to 5 minutes, the same as the setting in \cite{kishimoto2005search}. 
Note that in 5 minutes, about 150,000 simulations are performed using ELF and the one trained with FTL.
Results are shown in Table \ref{tab:rzone_experiments}.
Most notably, our program can solve 68 problems with RZS and FTL, 36 with RZS only (not FTL), and none without these methods (ELF).
In comparison, T-EXP can only solve 11 problems due to the large search space in the 19x19 board; it outperforms ELF (without RZS nor FTL), most likely due to its stronger domain knowledge.

The above L\&D problems, statistics, and training details can be accessed via the Github repository \cite{github_url}.

\section*{Discussion}
Our approach RZS is general in the sense that it is applicable to many other goal-achieving problems, in addition to Go and Hex.
A list of open issues worthy of further investigation is as follows.

\begin{itemize}
    \item 
Extend RZS to other goal-achieving problems for Go, such as seki and SSK, to solve more L\&D problems.
Incidentally, many experts expect a win for White for 7x7 kill-all Go. 
If so, the success of this work will make it more likely to solve 7x7 kill-all Go entirely. 
    \item 
Apply RZS to other goal-achieving problems for other games, such as Gomoku, Connect6 and Slither \cite{bonnet2015draws}, and even other domain of applications.
    \item 
Incorporate RZS into other search algorithms, such as proof-number search \cite{allis1994searching,nagai2002df} or alpha-beta search \cite{knuth1975analysis}.
    \item 
Finally, it is worth investigating whether zone patterns can act as features in pattern recognition to facilitate explainability in deep neural network training.
\end{itemize}

\section*{Acknowledgements}
This research is partially supported by the Ministry of Science and Technology (MOST) of Taiwan under Grant Numbers 110-2634-F-009-022, 110-2634-F-A49-004 and 110-2221-E-A49-067-MY3, and the computing resources are partially supported by National Center for High-performance Computing (NCHC) of Taiwan. The authors also thank Professor Martin M{\"u}ller and anonymous reviewers for their valuable comments.

\bibliography{aaai22.bib}

\section{Appendix}
\label{sec:appendix}

In this Appendix, we include the following materials.

\begin{itemize}
    \item Summarize how blocks on stones are defined to be {\em unconditionally alive (UCA)} for achieving safety in L\&D problems by \citet{benson1976life}. 
    \item Present zone dilation methods that expand zones to satisfy the three CR conditions and illustrate the cases where zone dilation is needed for L\&D problems. 
    \item Present a goal-achieving solver that combines RZS with depth-first search. 
    \item Introduce the game of Slither and illustrate its zone dilation using an example.
    \item Provide experiment details, including training settings, the collection of L\&D problems, and their run times. The L\&D problems are attached in SGF format in the supplementary materials. 
\end{itemize}

\subsection{Unconditionally Alive} 
\label{subsec:uca}

This subsection summarizes the definition of a block being unconditionally alive (UCA), as presented by \citet{benson1976life}. 
For a set of blocks to be UCA, each block needs to be associated with at least two {\em vital regions}, each of which is surrounded by blocks within that set.
For a region to be considered vital for a block, all empty grids within the region must be adjacent to that block.
For the position $p_b$ in Figure \ref{fig:benson_examples}, the (only) white block is UCA with the two vital regions at F2 and G1. 
UCA is also satisfied for the two white blocks in the position $p_c$, where the first vital region contains E2 and F2, and the other G1, with a similar situation in $p_d$. 
In contrast, UCA is not satisfied for the other four positions in Figure \ref{fig:benson_examples}, e.g., for $p_a$, the white block at D2 (one single stone) has no vital regions. 
Intuitively, since each block is associated with at least two vital regions, and each of these regions must have at least one empty grid, each block has at least two liberties, like the cases of $p_b$, $p_c$ and $p_d$ in Figure \ref{fig:benson_examples}.
Namely, each of these blocks can maintain at least two liberties even if the opponent is allowed unlimited consecutive moves. 
Thus, these blocks are safe or unconditional alive. 

\begin{figure}[th!]
    \centering
    \begin{subfigure}[t]{0.24\columnwidth}\includegraphics[width=\columnwidth]{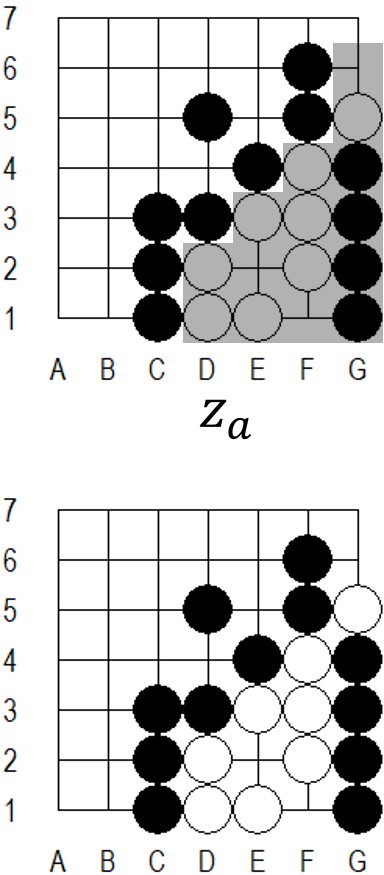}\caption{$p_a$}\end{subfigure}
    \begin{subfigure}[t]{0.24\columnwidth}\includegraphics[width=\columnwidth]{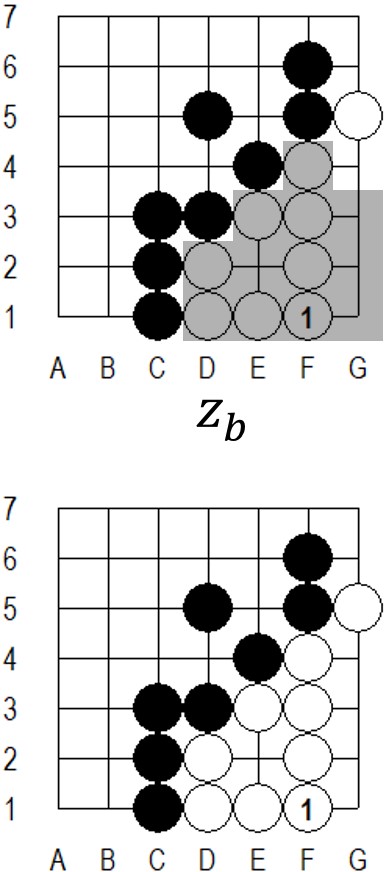}\caption{$p_b$}\end{subfigure}
    \begin{subfigure}[t]{0.24\columnwidth}\includegraphics[width=\columnwidth]{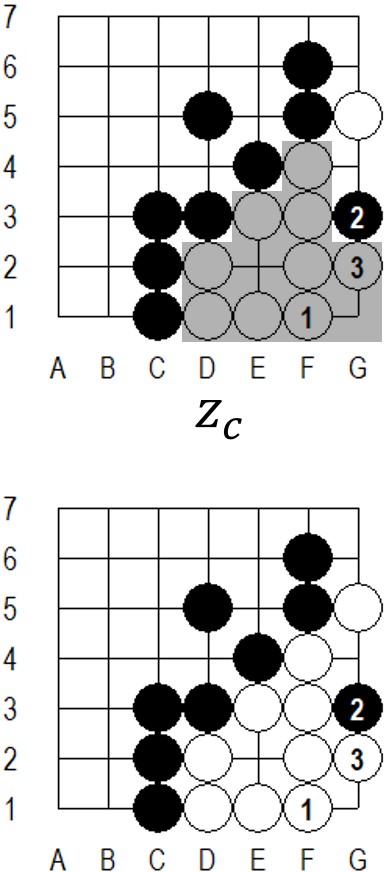}\caption{$p_c$}\end{subfigure}
    \begin{subfigure}[t]{0.24\columnwidth}\includegraphics[width=\columnwidth]{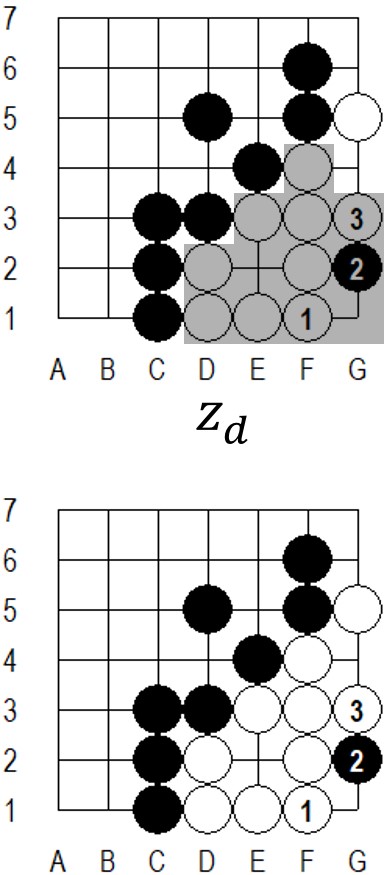}\caption{$p_d$}\end{subfigure}  
    \begin{subfigure}[t]{0.24\columnwidth}\includegraphics[width=\columnwidth]{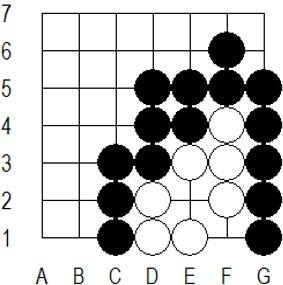}\caption{$p_e$}\end{subfigure}
    \caption{
    A dilation case for \ORplayer (White) with capturing blocks.
    For the position $p_a$ in (a), White is to play and wins as follows. 
    White plays at F1 to capture four black stones, leading to the position $p_b$. 
    Then, as in (c), for Black's move at G3, White replies G2 to achieve UCA and obtains an RZ $z_c$ as shaded. 
    Since G3 is outside of the RZ, it is a null move, which reduces the must-play region to G1 and G2 only. 
    It is trivial for White to reply at G2 for Black's move at G1. 
    For Black's move at G2, White replies at G3 to achieve UCA and obtains an RZ $z_d$ as shaded in (d). 
    Thus, we conclude that White wins for the position $p_b$ (Black to play).
    The union of RZs of the children of $p_b$ is $z_b$, as shaded in (b). This union satisfies the CR conditions, and thus is a valid RZ.
    Now, let us examine $p_a$ with White to play. 
    Let $z_u$ be the union of F1 (the move that was searched) and $z_b$, which is still $z_b$. 
    Unfortunately, in this case, $z_u$ can not serve as an RZ for $p_a$. As a counter-example, consider $p_e$, which shares the same pattern within $z_u$ as $p_a$. If $z_u$ is an RZ for $p_a$, for any position that shares the same pattern, say, $p_e$, White should be able to replay their winning strategy at F1. This is clearly not the case in $p_e$. 
    In order to replay the winning strategy (guaranteed by satisfying all the CR conditions), we need to dilate $z_u$ to $z_a$.
    }
    \label{fig:white_capture}
\end{figure}

\subsection{Zone Dilation} 
\label{subsec:dilation}

For all replayable solution trees rooted at an internal node $n$ (i.e. $p(n)$ is not UCA), their corresponding RZs $z$ are derived by dilating (expanding) some zone $z_u$ such that the resulting zone satisfies the three CR conditions so that replay is possible. Depending on whose turn it is to play at $n$, $z_u$ is defined as follows: 

\begin{description}
\item[OR-player (\ORplayer) to move]
$z_u$ is the union of the winning move $m$ and $z'$, where $z'$ is the RZ of the successor state of $n$ after playing $m$. 
\item[AND-player (\ANDplayer) to move]
$z_u$ is the union of RZs of their children. 
\end{description}

\begin{figure}[th!]
    \centering
    \begin{subfigure}[t]{0.32\columnwidth}\includegraphics[width=\columnwidth]{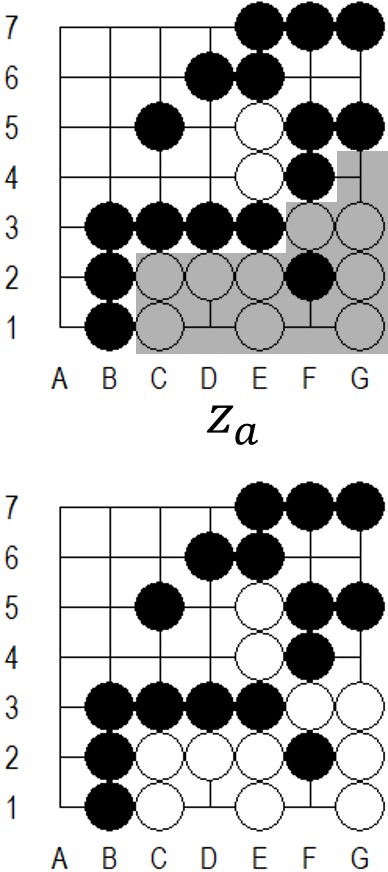}\caption{$p_a$}\end{subfigure}
    \begin{subfigure}[t]{0.32\columnwidth}\includegraphics[width=\columnwidth]{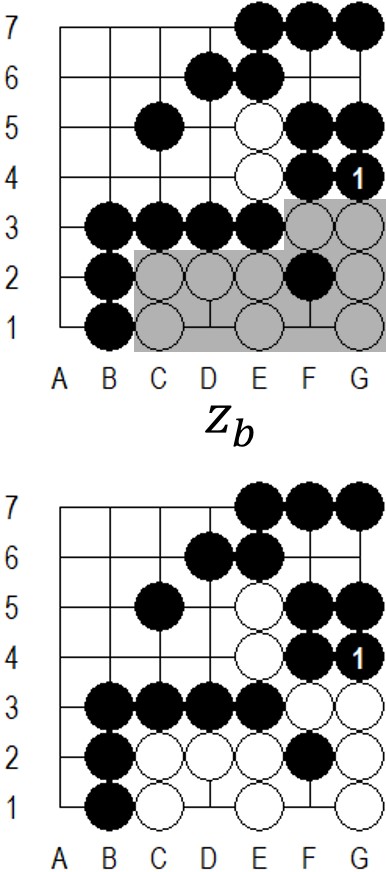}\caption{$p_b$}\end{subfigure}
    \begin{subfigure}[t]{0.32\columnwidth}\includegraphics[width=\columnwidth]{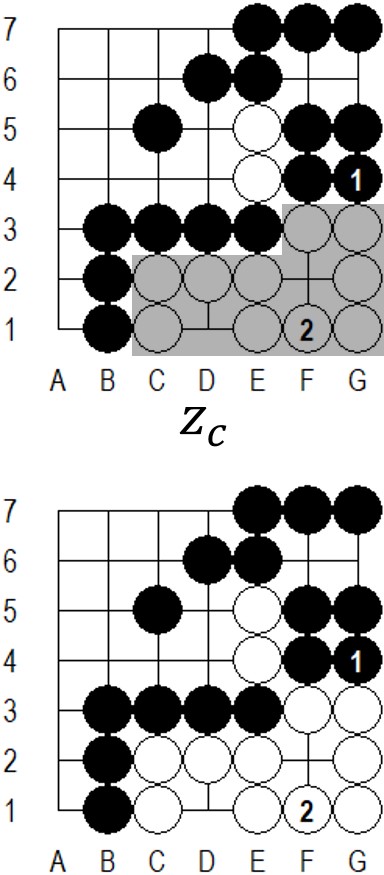}\caption{$p_c$}\end{subfigure}
    \begin{subfigure}[t]{0.32\columnwidth}\includegraphics[width=\columnwidth]{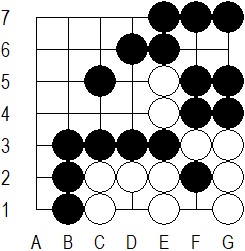}\caption{$p_d$}\end{subfigure}
    \begin{subfigure}[t]{0.32\columnwidth}\includegraphics[width=\columnwidth]{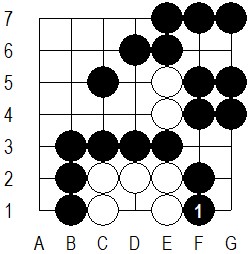}\caption{$p_e$}\end{subfigure}
    \caption{
    A dilation case for \ANDplayer (Black) to play.
    For $p_a$, where Black is to play, let Black play at G4 leading to $p_b$, and White reply at F1 leading to $p_c$. 
    Thus, $p_c$ achieves UCA, and the shaded grids in (c) is its RZ $z_c$.
    Then, for $p_b$ where White is to play and win at F1, F1 is replayable in the same zone, so its RZ $z_b$ is still $z_c$. 
    Since Black's move at G4 is outside of $z_b (=z_c)$, it is a null move and thus all moves outside $z_b$ are pruned. 
    This leads to a win for White in $p_a$ due to no more Black moves. 
    To derive the RZ of $p_a$, we first consider the union of the RZs for all Black moves (actually just one Black move at G4 here), still the same as $z_b$. 
    Unfortunately, $z_b$ cannot serve as an RZ for $p_a$ yet, since in $p_d$, which shares the same zone pattern, Black can directly capture the four White stones on the right by playing at F1, leading to $p_e$. 
    The RZ therefore needs to be dilated from $z_b$ to $z_a$ with one extra grid at G4. 
    }
    \label{fig:simple_dilate}
\end{figure}

\begin{figure}[th!]
    \centering
    \begin{subfigure}[t]{0.32\columnwidth}\includegraphics[width=\columnwidth]{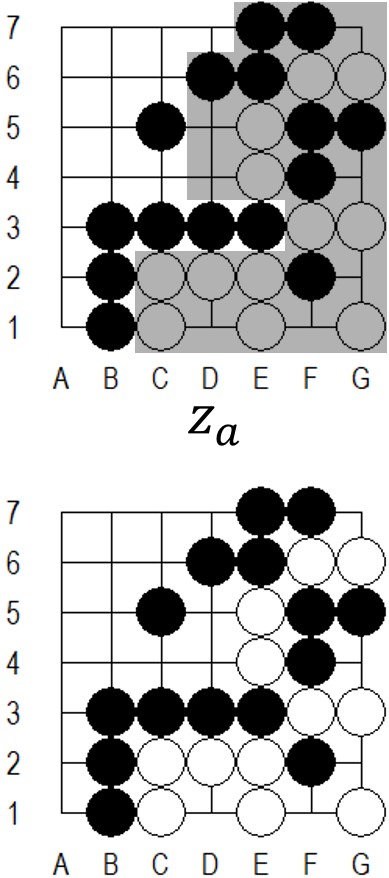}\caption{$p_a$}\end{subfigure}    
    \begin{subfigure}[t]{0.32\columnwidth}\includegraphics[width=\columnwidth]{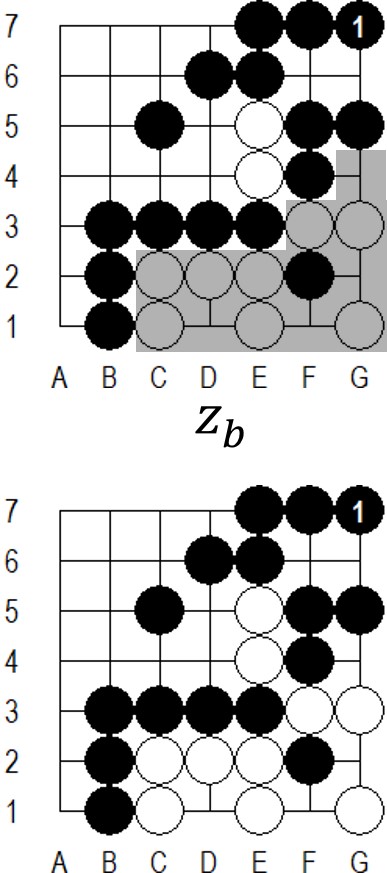}\caption{$p_b$}\end{subfigure}    
    \begin{subfigure}[t]{0.32\columnwidth}\includegraphics[width=\columnwidth]{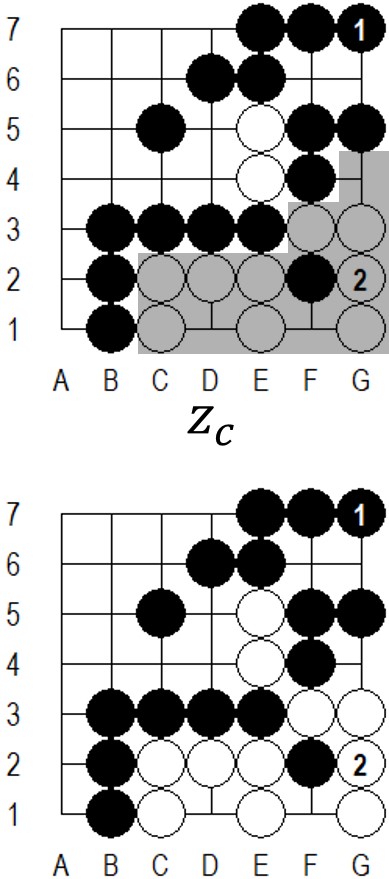}\caption{$p_c$}\end{subfigure}      
    \begin{subfigure}[t]{0.32\columnwidth}\includegraphics[width=\columnwidth]{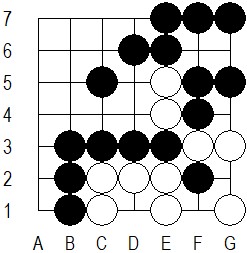}\caption{$p_d$}\end{subfigure}     
    \begin{subfigure}[t]{0.32\columnwidth}\includegraphics[width=\columnwidth]{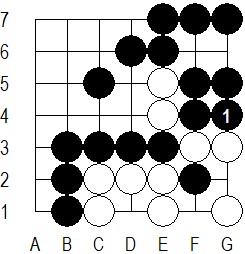}\caption{$p_e$}\end{subfigure}     
    \caption{
    A dilation case for \ANDplayer (Black) with suicide. 
    In the position $p_a$ in (a), Black cannot play at G4 since suicide is prohibited. Instead, Black tries G7 to capture two white stones as in (b). 
    White needs to respond at G2 to win in $p_c$.
    Since $p_c$ is identical to that in Figure \ref{fig:simple_dilate} (a), $p_c$ wins with the same zone, $z_c$ here. 
    For $p_b$ where it is White to play and win at G2, G2 is replayable in the same zone, so its RZ $z_b$ is still $z_c$. 
    Since Black's move at G7 is outside of $z_b (=z_c)$, all moves outside $z_b$ are pruned, leading to those moves inside $z_b$ for Black to move in (a). 
    In (a), since G4 and D1 are illegal, Black can only play F1 and G2, both of which Black can win with the same RZs (we omit the details for simplicity). 
    Consequently, the union of the RZs for all Black moves is still the same as $z_b$. 
    Unfortunately, $z_b$ cannot serve as an RZ for $p_a$, since there exists some position $p_d$ with the same zone pattern, except in this case Black can play at G4 to kill white stones as in (e). 
    So the zone needs to be dilated to $z_a$ as in (a). 
    }
    \label{fig:dilate_suicide}
\end{figure}

In practice, for most cases, $z$ remains the same as $z_u$, so long as $z_u$ already satisfies the CR conditions. 
For example, all cases in Figures \ref{fig:benson_examples} and \ref{fig:cross_example} do not require additional dilation. 
Let $z$ be initialized to $z_u$. 
Dilation needs to be performed to satisfy the CR conditions mainly in cases where suicide moves are prohibited and blocks are captured on the border of the RZ $z$, for the game of Go. 
Let us illustrate by the following three examples. 
The first is a dilation case for \ORplayer (White) with capturing blocks in Figure \ref{fig:white_capture},  
the second and the third are cases for \ANDplayer (Black) to play with capturing and suicides as in Figures \ref{fig:simple_dilate} and \ref{fig:dilate_suicide}.


To formulate a heuristic that dilates RZs while satisfying the CR conditions, let us review the following rules for Go.
\footnote{Note that we don't consider ko and SSK in this paper.}
\begin{description}
\item[GR-1] For all positions, all blocks must have at least one liberty, and a block without any liberties is removed from the board. 
\end{description}
Placing a stone on an unoccupied grid $g$ will lead to three kinds of situations.

\begin{description}
    \item[GR-2] One or more opponent's blocks without any more liberties are captured. This is a legal move. 
    \item[GR-3] No opponent blocks are captured, and one of the player's own blocks has no more liberties. This is a suicide, which is illegal in Go. 
    \item[GR-4] All blocks have at least one liberty. This is a legal move. 
\end{description}

Now, we give some new definitions related to $z$. 
Inside a zone $z$, a grid is called a $z$-grid, and a block is called a $z$-block. 
For a $z$-block, its liberties located inside $z$ are called $z$-liberties of the block.
A $z$-border is the set of $z$-grids which are adjacent to any of grids of $\bar{z}$.
For example, in Figure \ref{fig:benson_examples} (a), the $z$-border includes D1, D2, E3, F3 and G3, and D1 and E2 are the $z$-liberties of the white $z$-block at D2. 

In order to let $z$ satisfy the three CR conditions on $p(n)$, we propose some rules for dilation from a zone $z$. The first is DL-1 below, used to ensure that all $p_*$ with the same zone pattern, i.e., $\beta(p_*)\odot z=\beta(p(n))\odot z$, are legal at least inside the zone.

\begin{description}
    \item[DL-1] All $z$-blocks (both white and black) must have at least one $z$-liberty. 
    Assume that a $z$-block $b$ has no $z$-liberty. 
    Add one of $b$'s liberty into $z$. 
\end{description}

\subsubsection{Internal OR-Nodes}
Now, suppose that \ORplayer is to play and win at move $m$. 
In this case, the dilation only needs to satisfy CR-3. 
First, assume that \ORplayer wins without capturing any blocks. 
Then, simply apply DL-1 to $z$. 
It is trivial to see the CR-3 condition satisfied in this case.

Second, assume that \ORplayer wins by capturing one of \ANDplayer's block $b$. 
Then, add into $z$ the captured block $b$ and all \ORplayer's blocks surrounding $b$, and subsequently apply DL-1 to the resulting zone.
For example, in Figure \ref{fig:white_capture}, the captured black block (with four stones) and the white block at G5 surrounding the black block are included in $z_a$. 
In addition, DL-1 is then used to add the unoccupied grid at G6 into the RZ. 




\subsubsection{Internal AND-Nodes}
Now, suppose that \ANDplayer is to play and \ORplayer still wins on $p(n)$. 
In this case, we need to ensure both CR-1 and CR-2 conditions are satisfied. 
Let $p_*$ have the same zone pattern on $z$ as $p(n)$. 

For CR-1, all Black moves inside $z$ of $p_*$ must be legal for $p(n)$ with the same zone pattern. 
Equivalently, the following condition holds: For all illegal moves $m$ on the RZ $z$ in $p(n)$, $m$ are illegal in $p_*$. 
For the game of Go, since suicides are illegal, let us consider all suicides on $g$ inside the zone $z$ of $p(n)$.
Let $g$ be adjacent to \ANDplayer's block $b$, if any. 
To ensure that $g$ is also a suicide in $p_*$, add $b$ and all \ORplayer's blocks surrounding $b$ and $g$ into $z$. 
For example, in Figure \ref{fig:dilate_suicide} (a), Black move at G4 has the black block at (F4, F5, G5) have no liberty, so it is a suicide.
In this case, this Black block and two extra surrounding white blocks at (E4, E5) and (F6, G6) are added into $z$. 

For CR-2, all \ANDplayer's moves outside $z$ of $p_*$ cannot change the zone pattern of $p(n)$.
The only way to change the zone inside is to capture white $z$-blocks $b$ on the $z$-border. 
As GR-1, $b$ has at least one liberty in $p(n)$. 
So, let us consider the following two cases. 
In the case that $b$ has at least two liberties, add two of them into the zone $z$. 
For example, in Figure \ref{fig:dilate_suicide} (a), two liberties at D4 and D5 are added after the white block at (E4, E5) is added. 
Another example is in Figure \ref{fig:simple_dilate} (a), where the two liberties of the right white block (with four stones) need to be added.
In the case that $b$ has one and only one liberty, we add all the black blocks surrounding $b$. 
For example, in Figure \ref{fig:dilate_suicide} (a), after the white block at (F6, G6) is added, the only liberty of the block is at G7 which is in turn added and the surrounding black block at F7 should be added. 

Finally, we use DL-1 to dilate $z$. 
The above process is repeated until all are satisfied. 

\begin{algorithm}
\caption{AchieveGoal ($p$, $G$)}\label{alg:algo_achieve_goal}
\begin{algorithmic}[1]
\If {$Achieve(p, G)$} \Comment{Immediately achieves $G$}
    \State return ($win$, $z$), where $z$ is the RZ of $p$.
\EndIf
\If {$Fail(p, G)$} \Comment{Fails to achieve $G$}
    \State return $(fail, \emptyset)$
\EndIf
\If {$\pi(p)$ is \ORplayer}  \Comment{Internal OR-node}
    \While {M: more legal moves not searched yet}
        \State $(m, p') \gets$ nextmove($p$,$M$) \Comment{$p'$: next position}
        \State $(v, z') \gets$ AchieveGoal($p'$, $G$)
        \If {$v$ is $win$}  return $(win,$ dilate$(z' \cup \{m\}))$
        \EndIf
    \EndWhile
    \State return $(fail, \emptyset)$
\Else \Comment{Internal AND-node}
    \State $z \gets \emptyset$
    \State $M \gets$ all legal moves in $p$ \Comment{must-play region} 
    \While{more legal moves in $M$}
        \State $(m, p') \gets$ nextmove($p$,$M$) 
        \State $(v, z') \gets$ AchieveGoal($p'$, $G$)
        \If {$v$ is $win$} 
            \State $z \gets z \cup z'$
            \If {$m$ does not change pattern $\beta(p) \odot z'$ } 
            \State $M \gets M \cap z'$
            \Comment{$m$ is a null move}
            \Else ~ $M \gets M \setminus \{m\}$
            \EndIf
        \Else
            \State return $(fail, \emptyset)$
        \EndIf
    \EndWhile
    \State return $(win,$ dilate$(z)$) \Comment{$M$ is empty}
\EndIf
\end{algorithmic}
\end{algorithm}

\subsection{RZS Solver in Depth-First Search}

In this subsection, we present an RZS solver implemented within a depth-first search algorithm, which we call AchieveGoal. It is similar to the must-play-based Hex solver by \citet{hayward2009puzzling}. 
The algorithm AchieveGoal has inputs ($p$, $G$), where $p$ is the given position and $G$ the goal to achieve. $G$ is global and does not change during the process. The output is ($v$, $z$), where $v$ is either {\em win} if the goal is achieved, or {\em fail} otherwise. $z$ is the RZ, if $p$ is a win.
A heuristic function $selectmove(p,M)$ is used to choose a move from a set of moves $M$ on $p$.
When the algorithm returns a win, the function $dilate(z)$ also dilates the returning RZ to ensure that the three CR conditions are satisfied for all winning nodes. 
From Lemma \ref{lemma:consistent-replay}, the goal $G$ can be achieved for position $p$ with an RZ $z$, if AchieveGoal($p$, $G$) returns $(win, z)$. 

\subsection{RZS for the Game of Slither}
This subsection demonstrates that our RZS can even be applied to games where the stones move on the board as the game progresses. 

\begin{figure}[th]
    \centering
    \begin{subfigure}[t]{0.22\columnwidth}\includegraphics[width=\columnwidth]{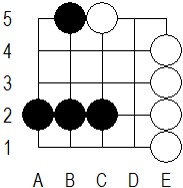}\caption{$p_a$}\end{subfigure}
    \begin{subfigure}[t]{0.22\columnwidth}\includegraphics[width=\columnwidth]{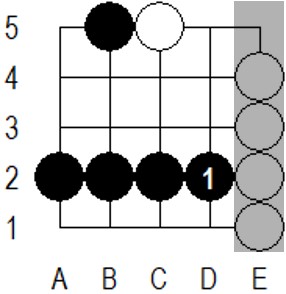}\caption{$p_b$}\end{subfigure}
    \begin{subfigure}[t]{0.22\columnwidth}\includegraphics[width=\columnwidth]{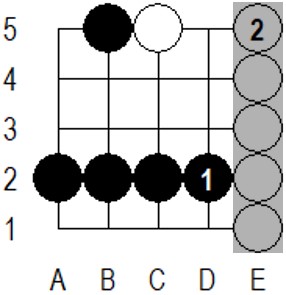}\caption{$p_c$}\end{subfigure}
    \\
    \begin{subfigure}[t]{0.22\columnwidth}\includegraphics[width=\columnwidth]{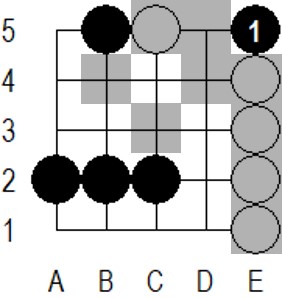}\caption{$p_d$}\end{subfigure}
    \begin{subfigure}[t]{0.22\columnwidth}\includegraphics[width=\columnwidth]{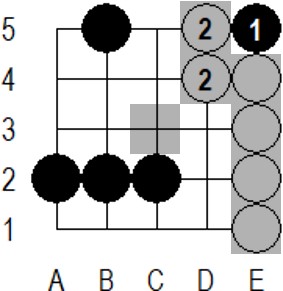}\caption{$p_e$}\end{subfigure}  
    \begin{subfigure}[t]{0.22\columnwidth}\includegraphics[width=\columnwidth]{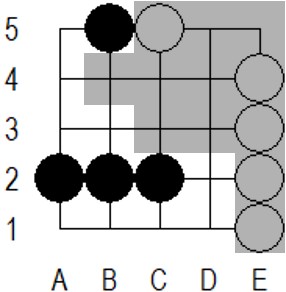}\caption{$p_a$ with $z_a$}\end{subfigure}        
    \caption{
    The position $p_a$ in (a) is for Black to play first and a win for White. 
    If Black plays at D2 as in (b), then White simply plays at E5 to connect the upper and lower sides to win with the RZ $z_c$, shaded in (c). 
    Since Black D2 is outside the RZ in (b), which is also $z_c$, it is a null move, and thus the must-play region is reduced to E5.
    Therefore, Black must play at E5 (no matter how Black moves stones in the first phase) to prevent White from winning directly as in (d). 
    For Black E5, White wins by moving the stone at C5 to D5, then placing a stone at D4 to win, leading to a win in (e). 
    Thus, position $p_a$ wins. 
    The RZ for the position in (e) is dilated to include C3 due to the diagonal rule.
    The RZ in (d) is dilated to include B4 (due to the diagonal rule with C5) and C5 itself. 
    The RZ for (a) is derived as shaded in (f) as follows.
    Take the union of the above two RZs in both (b) and (d). Then dilate it by including C3, C4, C5, and D3, since Black can move any one of them to D4 and D5 to prevent White's winning move as in (e). 
}
    \label{fig:slither_examples}
\end{figure}

Slither is also a connection game play on an $n \times n$ square board, in which stones can be placed on unoccupied grids and moved. 
It has been shown that the game is PSPACE-complete and cannot end in a draw \cite{bonnet2015draws}.
Like the game of Hex, the first player who connects its own sides (e.g., up/down for White) wins.
A move in Slither is composed of two phases. 
The first is an optional relocation which allows the player to shift an existing stone of their color to an adjacent or diagonal empty grid. 
The second is to place a stone on another empty grid. 
The resulting position cannot have any two diagonal grids occupied by the player's stones which do not have an orthogonally-adjacent stone, as illustrated as follows. 
For example, it is illegal for White to add a single stone at D5 into the position in Figure \ref{fig:slither_examples} (a), due to the diagonal (D5, E4). 
However, the diagonal (D5, E4) is legal in (e) since D4 connects to both via a direct adjacency. 
This restriction is called the diagonal rule  \cite{bonnet2015draws}.
Figure \ref{fig:slither_examples} (a) illustrates an winning case for White, and the zone dilation for the RZ.
More zone dilation methods for Slither are beyond the scope of this paper.

\subsection{Experiment Details}
The following lists the training settings for two AlphaZero programs with and without the FTL method in 7x7 kill-all Go.
The network architecture consists of 5 residual blocks and 64 filters.
We generate a total of 2,500,000 self-play games in training with 400 MCTS simulations for each move.
The learning rate is fixed to 0.02 for the first 1,500,000 self-play games, then reduced to 0.002 for the remainder; the komi is set to 48 during training.

\begin{figure}[th]
    \centering
    \begin{subfigure}[t]{0.32\columnwidth}\includegraphics[width=\columnwidth]{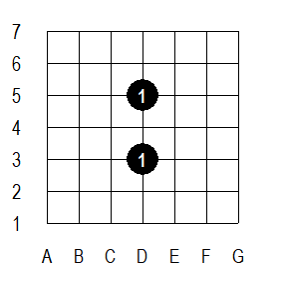}\caption{One-point jump.}\end{subfigure}
    \begin{subfigure}[t]{0.32\columnwidth}\includegraphics[width=\columnwidth]{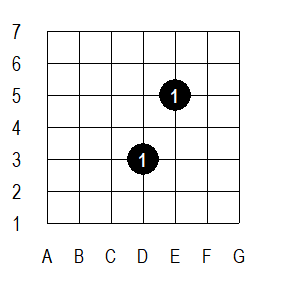}\caption{Knight's move.}\end{subfigure}
    \begin{subfigure}[t]{0.32\columnwidth}\includegraphics[width=\columnwidth]{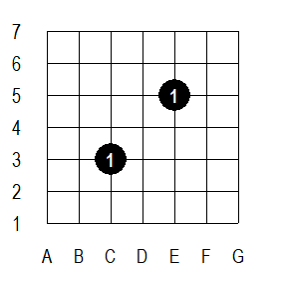}\caption{Diagonal jump.}\end{subfigure}
    \caption{Openings for 7x7 kill-all Go.}
    \label{fig:77opening}
\end{figure}

\begin{figure*}[t]
    \centering
	\begin{subfigure}{0.45\columnwidth}\includegraphics[width=\columnwidth]{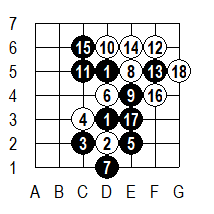}\caption{1. Black's turn.}\end{subfigure}
	\begin{subfigure}{0.45\columnwidth}\includegraphics[width=\columnwidth]{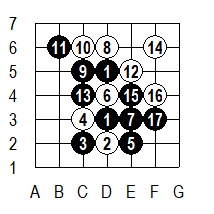}\caption{2. White's turn.}\end{subfigure}
	\begin{subfigure}{0.45\columnwidth}\includegraphics[width=\columnwidth]{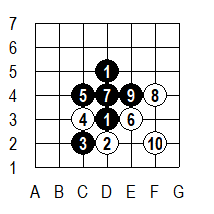}\caption{3. Black's turn.}\end{subfigure}
	\begin{subfigure}{0.45\columnwidth}\includegraphics[width=\columnwidth]{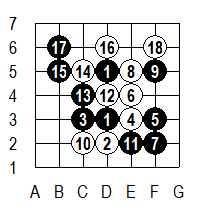}\caption{4. Black's turn.}\end{subfigure}

	\begin{subfigure}{0.45\columnwidth}\includegraphics[width=\columnwidth]{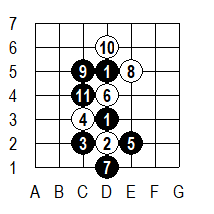}\caption{5. White's turn.}\end{subfigure}
	\begin{subfigure}{0.45\columnwidth}\includegraphics[width=\columnwidth]{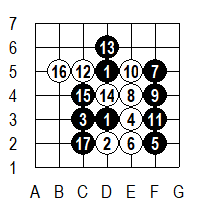}\caption{6. White's turn.}\end{subfigure}
	\begin{subfigure}{0.45\columnwidth}\includegraphics[width=\columnwidth]{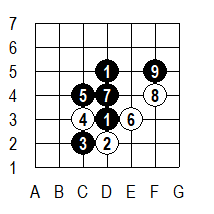}\caption{7. White's turn.}\end{subfigure}
	\begin{subfigure}{0.45\columnwidth}\includegraphics[width=\columnwidth]{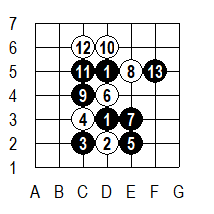}\caption{8. White's turn.}\end{subfigure}

	\begin{subfigure}{0.45\columnwidth}\includegraphics[width=\columnwidth]{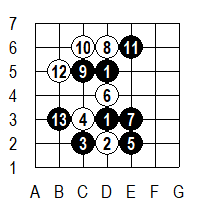}\caption{9. White's turn.}\end{subfigure}
	\begin{subfigure}{0.45\columnwidth}\includegraphics[width=\columnwidth]{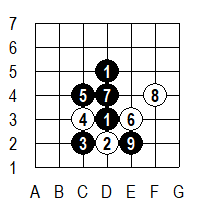}\caption{10. White's turn.}\end{subfigure}
	\begin{subfigure}{0.45\columnwidth}\includegraphics[width=\columnwidth]{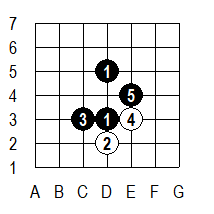}\caption{11. White's turn.}\end{subfigure}
	\begin{subfigure}{0.45\columnwidth}\includegraphics[width=\columnwidth]{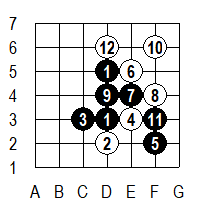}\caption{12. Black's turn.}\end{subfigure}

	\begin{subfigure}{0.45\columnwidth}\includegraphics[width=\columnwidth]{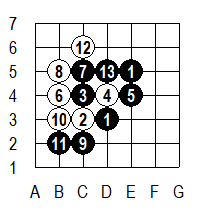}\caption{13. White's turn.}\end{subfigure}
	\begin{subfigure}{0.45\columnwidth}\includegraphics[width=\columnwidth]{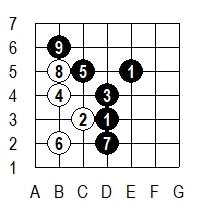}\caption{14. White's turn.}\end{subfigure}
	\begin{subfigure}{0.45\columnwidth}\includegraphics[width=\columnwidth]{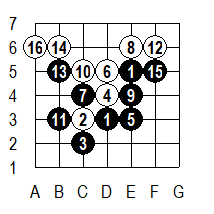}\caption{15. Black's turn.}\end{subfigure}
	\begin{subfigure}{0.45\columnwidth}\includegraphics[width=\columnwidth]{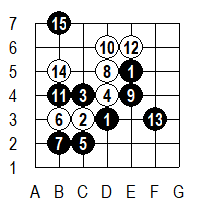}\caption{16. White's turn.}\end{subfigure}

	\begin{subfigure}{0.45\columnwidth}\includegraphics[width=\columnwidth]{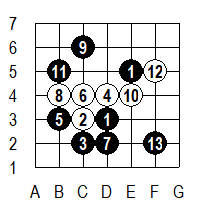}\caption{17. White's turn.}\end{subfigure}
	\begin{subfigure}{0.45\columnwidth}\includegraphics[width=\columnwidth]{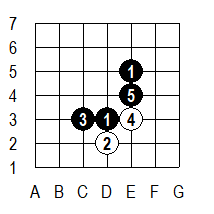}\caption{18. White's turn.}\end{subfigure}
	\begin{subfigure}{0.45\columnwidth}\includegraphics[width=\columnwidth]{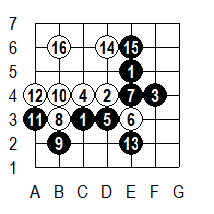}\caption{19. Black's turn.}\end{subfigure}
	\begin{subfigure}{0.45\columnwidth}\includegraphics[width=\columnwidth]{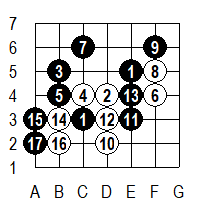}\caption{20. White's turn.}\end{subfigure}
    \caption{20 7x7 kill-all tsumego problems.}
    \label{fig:77_20_tsumego}
\end{figure*}

We select 20 L\&D problems for 7x7 kill-all Go from three different Black openings: the one-point jump opening, knight's move opening, and diagonal jump opening, as shown in Figure \ref{fig:77opening} (a), (b), and (c) respectively.
Figure \ref{fig:77_20_tsumego} lists these 20 problems in three groups, 12 from the one-point jump opening, 6 from the knight's move opening, and 2 from the diagonal jump opening. In each group, problems are sorted from simple to complex, in terms of simulation count for RZS with FTL. 
Particularly, the two problems 11 and 18, as in Figure \ref{fig:77_20_tsumego} (k) and (r) respectively, are commonly played as joseki (openings) in 19x19.

Table \ref{tab:7x7_killAll_detail} lists the simulation counts (or node counts) required to solve each problem for all methods. 
In the table, dashes indicate that the problem cannot be solved within the given computation budget.

\begin{figure}[h]
    \centering
	\begin{subfigure}{0.65\columnwidth}\includegraphics[width=\columnwidth]{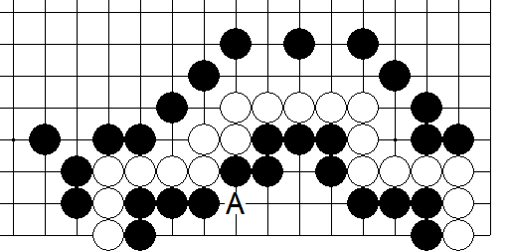}\caption{Original L\&D problem. }\end{subfigure}
	\begin{subfigure}{0.9\columnwidth}\includegraphics[width=\columnwidth]{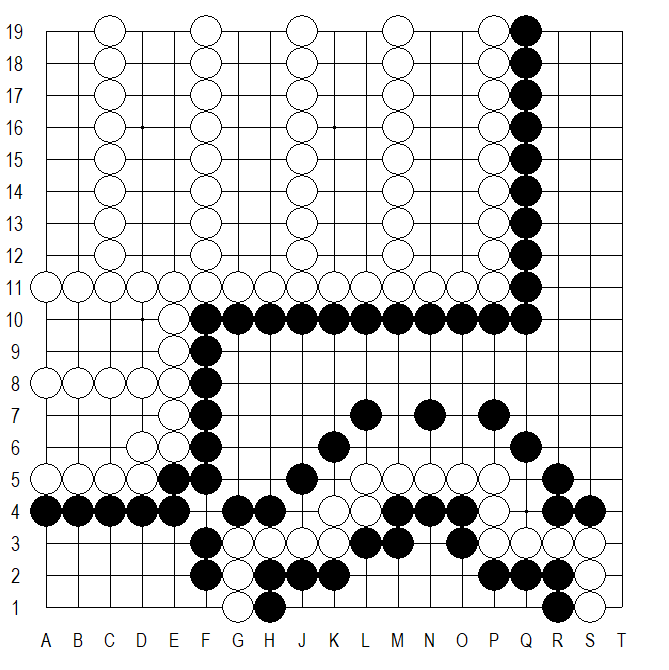}\caption{A modified 19x19 L\&D problem from (a).}\end{subfigure}
    \caption{A 19x19 L\&D problem from Cho's book, volume 2, page 139. The problem is White to play for life, and the correct answer is to play at A.}
    \label{fig:19x19_masked}
\end{figure}

Table \ref{tab:19x19_tsumego_detail} lists the experiment results for 19x19 Go L\&D problems, where the columns of volume and page number indicate the problem corresponding to the problem in that page of the volume of the book \cite{cho1987life}.
To satisfy the requirements of each solver, we modified the problems by filling some stones.
One of the modified L\&D problems is shown in Figure \ref{fig:19x19_masked} (b), where the original problem is shown in Figure \ref{fig:19x19_masked} (a).

The above L\&D problems are attached in SGF format in the supplementary material (in the directory "tsumego"). 

\begin{table*}[bp]
    \centering
    \begin{tabular}{|c|r|r|r|r|}
    \hline
    ID	&	No-RZS	&	RZS	&	FTL(w/o RZS)	&	FTL(w/ RZS)  	\\	\hline
    1	&	-	&	255,314	&	76,780	&	\textbf{325}	\\	\hline
    2	&	-	&	204,355	&	428,239	&	\textbf{4,958}	\\	\hline
    3	&	-	&	-	&	-	&	\textbf{217,461}	\\	\hline
    4	&	-	&	-	&	-	&	\textbf{106,430}	\\	\hline
    5	&	-	&	-	&	-	&	\textbf{204,691}	\\	\hline
    6	&	-	&	-	&	-	&	\textbf{177,231}	\\	\hline
    7	&	-	&	-	&	-	&	\textbf{173,056}	\\	\hline
    8	&	-	&	-	&	-	&	\textbf{164,633}	\\	\hline
    9	&	-	&	-	&	-	&	\textbf{180,901}	\\	\hline
    10	&	-	&	-	&	-	&	\textbf{335,878}	\\	\hline
    11	&	-	&	-	&	-	&	\textbf{449,791}	\\	\hline
    12	&	-	&	-	&	-	&	\textbf{365,809}	\\	\hline
    13	&	-	&	23,998	&	127,910	&	\textbf{1,129}	\\	\hline
    14	&	-	&	220,988	&	-	&	\textbf{915}	\\	\hline
    15	&	-	&	245,360	&	-	&	\textbf{2,402}	\\	\hline
    16	&	-	&	-	&	-	&	\textbf{18,279}	\\	\hline
    17	&	-	&	-	&	-	&	\textbf{319,276}	\\	\hline
    18	&	-	&	-	&	-	&	\textbf{423,564}	\\	\hline
    19	&	-	&	-	&	-	&	\textbf{7,067}	\\	\hline
    20	&	-	&	-	&	-	&	\textbf{13,137}	\\	\hline
    \end{tabular}
    \caption{Node counts on solving 7x7 kill-all Go problems. The bold numbers show the lowest number of search nodes to prove the problem among all methods.}
    \label{tab:7x7_killAll_detail}
\end{table*}

\begin{table*}[tbp]
    \centering
    \begin{tabular}{|c|r|r|r|r|r|r|r|r|}
    \hline
    ID	&	Volume	&	Page No.	&	ELF(w/o RZS)	&	ELF(w/ RZS)	&	FTL(w/o RZS)	&	FTL(w/ RZS)	&	T-EXP	    \\	\hline
	1	&	1	&	31	&		-		&		12,804		&		42,472		&	\textbf{2,596}	&		-		\\	\hline
	2	&	1	&	37	&		-		&	\textbf{32,518}	&		-		&		38,275		&		-		\\	\hline
	3	&	1	&	40	&		-		&		40,002		&		-		&	\textbf{9,375}	&		-		\\	\hline
	4	&	1	&	42	&		-		&		8,419		&		32,980		&	\textbf{416}	&		1,037		\\	\hline
	5	&	1	&	46	&		-		&	\textbf{2,434}	&		161,040		&		65,687		&		1,036,877		\\	\hline
	6	&	1	&	57	&		-		&	\textbf{3,499}	&		-		&		3,878		&		-		\\	\hline
	7	&	1	&	60	&		-		&		-		&		-		&	\textbf{27,974}	&		-		\\	\hline
	8	&	1	&	68	&		-		&		-		&		-		&	\textbf{1,960}	&		-		\\	\hline
	9	&	1	&	74	&		-		&		2,911		&		-		&	\textbf{1,547}	&		1,158,037		\\	\hline
	10	&	1	&	80	&		-		&		-		&		-		&	\textbf{39,758}	&		-		\\	\hline
	11	&	1	&	84	&		-		&		-		&		-		&	\textbf{11,422}	&		-		\\	\hline
	12	&	1	&	88	&		-		&		-		&		-		&		-		&		-		\\	\hline
	13	&	1	&	90	&		-		&		-		&		-		&		-		&		-		\\	\hline
	14	&	1	&	98	&		-		&		-		&		-		&		-		&		-		\\	\hline
	15	&	1	&	101	&		-		&	\textbf{17,165}	&		-		&		27,103		&		-		\\	\hline
	16	&	1	&	104	&		-		&		33,177		&		-		&	\textbf{6,071}	&		-		\\	\hline
	17	&	1	&	132	&		-		&		-		&		-		&	\textbf{65,713}	&		-		\\	\hline
	18	&	1	&	135	&		-		&		-		&		-		&	\textbf{212,209}	&		-		\\	\hline
	19	&	1	&	139	&		-		&		-		&		-		&	\textbf{15,964}	&		-		\\	\hline
	20	&	1	&	152	&		-		&		-		&		-		&		-		&		-		\\	\hline
	21	&	1	&	156	&		-		&		6,540		&		-		&	\textbf{2,031}	&		-		\\	\hline
	22	&	1	&	174	&		-		&		-		&		-		&	\textbf{30,897}	&		-		\\	\hline
	23	&	1	&	176	&		-		&		-		&		-		&	\textbf{26,562}	&		-		\\	\hline
	24	&	1	&	186	&		-		&		-		&		-		&		-		&		-		\\	\hline
	25	&	1	&	189	&		-		&		50,399		&		-		&	\textbf{25,147}	&		-		\\	\hline
	26	&	1	&	195	&		-		&		8,698		&		-		&	\textbf{7,055}	&		-		\\	\hline
	27	&	1	&	196	&		-		&		-		&		-		&	\textbf{23,738}	&		648,529		\\	\hline
	28	&	1	&	197	&		-		&		-		&		-		&	\textbf{44,385}	&		-		\\	\hline
	29	&	1	&	198	&		-		&	\textbf{8,080}	&		-		&		11,921		&		-		\\	\hline
	30	&	1	&	222	&		-		&		-		&		-		&		-		&		-		\\	\hline
	31	&	1	&	243	&		-		&	\textbf{776}	&		33,768		&		2,108		&		2,811		\\	\hline
	32	&	1	&	244	&		-		&	\textbf{8,223}	&		-		&		48,851		&		87,168		\\	\hline
	33	&	1	&	252	&		-		&	\textbf{4,053}	&		-		&		5,171		&		-		\\	\hline
	34	&	1	&	258	&		-		&		4,657		&		37,219		&	\textbf{434}	&		556		\\	\hline
	35	&	1	&	262	&		-		&		-		&		89,912		&	\textbf{739}	&		1,040		\\	\hline
	36	&	1	&	264	&		-		&		16,178		&		-		&	\textbf{6,062}	&		-		\\	\hline
	37	&	1	&	268	&		-		&	\textbf{5,466}	&		-		&		16,438		&		-		\\	\hline
	38	&	1	&	269	&		-		&	\textbf{3,693}	&		-		&		5,625		&		-		\\	\hline
	39	&	1	&	270	&		-		&	\textbf{4,956}	&		-		&		22,998		&		-		\\	\hline
	40	&	1	&	272	&		-		&	\textbf{1,539}	&		-		&		22,058		&		131,050		\\	\hline
	41	&	1	&	279	&		-		&	\textbf{8,463}	&		-		&		13,240		&		345,351		\\	\hline
	42	&	1	&	288	&		-		&		-		&		-		&	\textbf{35,664}	&		-		\\	\hline
	43	&	1	&	298	&		-		&		-		&		-		&	\textbf{10,804}	&		-		\\	\hline
	44	&	1	&	299	&		-		&		45,906		&		-		&	\textbf{32,555}	&		-		\\	\hline
	45	&	1	&	300	&		-		&	\textbf{25,915}	&		-		&		46,271		&		-		\\	\hline
	46	&	1	&	301	&		-		&		-		&		-		&	\textbf{18,811}	&		-		\\	\hline
	47	&	1	&	302	&		-		&		-		&		-		&		-		&		-		\\	\hline
	48	&	1	&	318	&		-		&		-		&		-		&	\textbf{190,704}	&		-		\\	\hline
	49	&	1	&	323	&		-		&		-		&		-		&	\textbf{12,544}	&		-		\\	\hline
	50	&	1	&	324	&		-		&		34,065		&		-		&	\textbf{26,933}	&		-		\\	\hline
    \end{tabular}
\end{table*}

\begin{table*}[tbp]
    \centering
    \begin{tabular}{|c|r|r|r|r|r|r|r|r|}
    \hline
    ID	&	Volume	&	Page No.	&	ELF(w/o RZ)	&	ELF(w/ RZ)	&	FTL(w/o RZ)	&	FTL(w/ RZ)	&	T-EXP	\\	\hline
	51	&	2	&	120	&		-		&		75,016		&		-		&	\textbf{28,647}	&		-		\\	\hline
	52	&	2	&	124	&		-		&		-		&		-		&		-		&		-		\\	\hline
	53	&	2	&	125	&		-		&		-		&		-		&	\textbf{180,507}	&		-		\\	\hline
	54	&	2	&	126	&		-		&		-		&		-		&		-		&		-		\\	\hline
	55	&	2	&	127	&		-		&		-		&		-		&	\textbf{58,900}	&		-		\\	\hline
	56	&	2	&	129	&		-		&		-		&		-		&		-		&		-		\\	\hline
	57	&	2	&	132	&		-		&		-		&		-		&		-		&		-		\\	\hline
	58	&	2	&	133	&		-		&		7,424		&		-		&	\textbf{5,287}	&		-		\\	\hline
	59	&	2	&	138	&		-		&		-		&		-		&		-		&		-		\\	\hline
	60	&	2	&	139	&		-		&		-		&		-		&	\textbf{195,445}	&		-		\\	\hline
	61	&	2	&	142	&		-		&		-		&		-		&		-		&		-		\\	\hline
	62	&	2	&	144	&		-		&		-		&		-		&		-		&		-		\\	\hline
	63	&	2	&	146	&		-		&		-		&		-		&		-		&		-		\\	\hline
	64	&	2	&	147	&		-		&		-		&		-		&	\textbf{251,136}	&		-		\\	\hline
	65	&	2	&	152	&		-		&		-		&		-		&		-		&		-		\\	\hline
	66	&	2	&	156	&		-		&		-		&		-		&	\textbf{59,898}	&		-		\\	\hline
	67	&	2	&	158	&		-		&		-		&		-		&		-		&		-		\\	\hline
	68	&	2	&	162	&		-		&	\textbf{95,163}	&		-		&		-		&		-		\\	\hline
	69	&	2	&	164	&		-		&		-		&		-		&		-		&		-		\\	\hline
	70	&	2	&	166	&		-		&		-		&		-		&		-		&		-		\\	\hline
	71	&	2	&	245	&		-		&		-		&		-		&		-		&		-		\\	\hline
	72	&	2	&	246	&		-		&		-		&		-		&	\textbf{5,448}	&		-		\\	\hline
	73	&	2	&	248	&		-		&		-		&		-		&		-		&		-		\\	\hline
	74	&	2	&	252	&		-		&		90,000		&		-		&	\textbf{7,483}	&		-		\\	\hline
	75	&	2	&	253	&		-		&		-		&		-		&		-		&		-		\\	\hline
	76	&	2	&	254	&		-		&		-		&		-		&		-		&		-		\\	\hline
	77	&	2	&	257	&		-		&		-		&		-		&		-		&		-		\\	\hline
	78	&	2	&	260	&		-		&		-		&		-		&		-		&		-		\\	\hline
	79	&	2	&	262	&		-		&		-		&		-		&		-		&		-		\\	\hline
	80	&	2	&	274	&		-		&		-		&		-		&		-		&		-		\\	\hline
	81	&	2	&	326	&		-		&		-		&		-		&		-		&		-		\\	\hline
	82	&	2	&	328	&		-		&		-		&		-		&	\textbf{43,247}	&		-		\\	\hline
	83	&	2	&	329	&		-		&		13,668		&		-		&	\textbf{2,009}	&		-		\\	\hline
	84	&	2	&	330	&		-		&		41,389		&		-		&	\textbf{22,957}	&		-		\\	\hline
	85	&	2	&	331	&		-		&		-		&		-		&		-		&		-		\\	\hline
	86	&	2	&	332	&		-		&		-		&		-		&		-		&		-		\\	\hline
	87	&	2	&	333	&		-		&		33,092		&		-		&	\textbf{9,701}	&		-		\\	\hline
	88	&	2	&	334	&		-		&		-		&		-		&		-		&		-		\\	\hline
	89	&	2	&	336	&		-		&		-		&		-		&	\textbf{15,996}	&		-		\\	\hline
	90	&	2	&	337	&		-		&		12,582		&		-		&	\textbf{10,544}	&		-		\\	\hline
	91	&	2	&	338	&		-		&		-		&		-		&	\textbf{33,088}	&		-		\\	\hline
	92	&	2	&	340	&		-		&		-		&		-		&		-		&		-		\\	\hline
	93	&	2	&	341	&		-		&		-		&		-		&	\textbf{16,056}	&		-		\\	\hline
	94	&	2	&	342	&		-		&		-		&		-		&	\textbf{173,118}	&		-		\\	\hline
	95	&	2	&	343	&		-		&		103,057		&		-		&	\textbf{12,210}	&		-		\\	\hline
	96	&	2	&	344	&		-		&		-		&		-		&	\textbf{37,943}	&		-		\\	\hline
	97	&	2	&	345	&		-		&	\textbf{7,424}	&		-		&		27,276		&		8,385		\\	\hline
	98	&	2	&	351	&		-		&		-		&		-		&		-		&		-		\\	\hline
	99	&	2	&	352	&		-		&		-		&		-		&	\textbf{23,855}	&		-		\\	\hline
	100	&	2	&	353	&		-		&		-		&		-		&		-		&		-		\\	\hline
    \end{tabular}
\end{table*}

\begin{table*}[tbp]
    \centering
    \begin{tabular}{|c|r|r|r|r|r|r|r|r|}
    \hline
    ID	&	Volume	&	Page No.	&	ELF(w/o RZ)	&	ELF(w/ RZ)	&	FTL(w/o RZ)	&	FTL(w/ RZ)	&	T-EXP	\\	\hline
	101	&	2	&	355	&		-		&		-		&		-		&		-		&		-		\\	\hline
	102	&	2	&	359	&		-		&		-		&		-		&		-		&		-		\\	\hline
	103	&	2	&	363	&		-		&		-		&		-		&		-		&		-		\\	\hline
	104	&	2	&	364	&		-		&		-		&		-		&	\textbf{17,458}	&		-		\\	\hline
	105	&	2	&	366	&		-		&		-		&		-		&	\textbf{59,295}	&		-		\\	\hline
	106	&	2	&	372	&		-		&		-		&		-		&	\textbf{72,562}	&		-		\\	\hline
    \end{tabular}
    \caption{Node counts on solving 19x19 L\&D problems. The bold numbers indicate the lowest number of search nodes to prove the problem among all methods.}
    \label{tab:19x19_tsumego_detail}
\end{table*}

\end{document}